\documentclass{article}

\usepackage{PRIMEarxiv}

\usepackage[utf8]{inputenc} 
\usepackage[T1]{fontenc}    
\usepackage{hyperref}       
\usepackage{url}            
\usepackage{booktabs}       
\usepackage{amsfonts}       
\usepackage{nicefrac}       
\usepackage{microtype}      
\usepackage{lipsum}
\usepackage{fancyhdr}       
\usepackage{graphicx}       
\graphicspath{{media/}}     
\usepackage{verbatim}       

\usepackage{tikz}           
\usetikzlibrary{shapes.geometric, arrows, positioning}

\usepackage[british]{babel}
\usepackage{hhline}
\usepackage{multirow}
\usepackage{subcaption}
\usepackage{colortbl}
\usepackage{makecell} 
\usepackage{colortbl} 
\usepackage{makecell} 
\usepackage{amsmath} 
\usepackage{array}
\usepackage{caption}  

\usepackage{float}  

\newcommand{\beginsupplement}{%
        \setcounter{table}{0}
        \renewcommand{\thetable}{S\arabic{table}}%
        \setcounter{figure}{0}
        \renewcommand{\thefigure}{S\arabic{figure}}%
        \setcounter{lstlisting}{0}
        \renewcommand{\thelstlisting}{S\arabic{lstlisting}} 
        \setcounter{section}{0}  
\renewcommand{\thesection}{S\arabic{section}}  
     }

\usepackage{booktabs}
\usepackage{siunitx}
\usepackage{threeparttable}

\usepackage{listings}
\usepackage{xcolor}

\definecolor{pythonblue}{RGB}{32,74,135}
\definecolor{pythonstring}{RGB}{164,0,0}
\definecolor{pythoncomment}{RGB}{96,96,96}

\lstdefinestyle{pythonstyle}{
    language=Python,
    basicstyle=\ttfamily\small,
    commentstyle=\color{pythoncomment},
    stringstyle=\color{pythonstring},
    keywordstyle=\color{pythonblue},
    showstringspaces=false,
    numbers=left,
    numberstyle=\tiny\color{pythoncomment},
    frame=single,
    rulecolor=\color{black},
    backgroundcolor=\color{white},
    breaklines=true,
    breakatwhitespace=true,
    tabsize=4
}

\definecolor{backcolour}{rgb}{0.95,0.95,0.92}
\definecolor{commentcolour}{rgb}{0.5,0.5,0.5}
\definecolor{stringcolour}{rgb}{0.58,0,0.82}
\definecolor{keywordcolour}{rgb}{0,0,0.6}
\definecolor{numbercolour}{rgb}{0.1,0.1,0.5}

\lstdefinelanguage{json}{
    basicstyle=\ttfamily\small,
    keywordstyle=\color{blue},
    stringstyle=\color{purple},
    commentstyle=\color{gray},
    morestring=[b]",
    morecomment=[l]{//},
    morecomment=[s]{/*}{*/},
    showstringspaces=false,
    showspaces=false,
    showtabs=false,
    tabsize=2,
    breaklines=true,
    frame=single
}

\definecolor{backcolour}{rgb}{0.95,0.95,0.92}
\definecolor{commentcolour}{rgb}{0.5,0.5,0.5}
\definecolor{stringcolour}{rgb}{0.58,0,0.82}
\definecolor{keywordcolour}{rgb}{0,0,0.6}
\definecolor{numbercolour}{rgb}{0.1,0.1,0.5}

\title{VISION: A Modular AI Assistant for Natural Human-Instrument Interaction at Scientific User Facilities   
}

\author{
  Shray Mathur \\
  Center for Functional Nanomaterials\\
  Brookhaven National Laboratory \\
  Upton, NY 11973, USA\\
   \And
  Noah van der Vleuten\\
  Center for Functional Nanomaterials\\
  Brookhaven National Laboratory \\
  Upton, NY 11973, USA\\
  \And
  Kevin G. Yager \\
  Center for Functional Nanomaterials\\
  Brookhaven National Laboratory \\
  Upton, NY 11973, USA\\
  \texttt{kyager@bnl.gov} \\
  \And
  Esther Tsai\\
  Center for Functional Nanomaterials\\
  Brookhaven National Laboratory \\
  Upton, NY 11973, USA\\
  \texttt{etsai@bnl.gov} \\
}

\begin{document}
\maketitle

\begin{abstract}

Scientific user facilities, such as synchrotron beamlines, are equipped with a wide array of hardware and software tools that require a codebase for human-computer-interaction.
This often necessitates developers to be involved to establish connection between users/researchers and the complex instrumentation.
The advent of generative AI presents an opportunity to bridge this knowledge gap, enabling seamless communication and efficient experimental workflows. 
Here we present a modular architecture for the {\underline{Vi}}rtual
 {\underline{S}}cientific Compan{\underline{ion}} (VISION) 
by assembling multiple AI-enabled cognitive blocks that each scaffolds large language models (LLMs) for a specialized task.
With VISION, we performed LLM-based operation on the beamline workstation with low latency and demonstrated the first voice-controlled experiment at an X-ray scattering beamline.
The modular and scalable architecture allows for easy adaptation to new instrument and capabilities.
Development on natural language-based scientific experimentation is a building block for an impending future where a science exocortex---a synthetic extension to the cognition of scientists---may radically transform scientific practice and discovery.


\end{abstract}

\section{Introduction}

Research in basic energy science often aims to reveal the underlying mechanisms of the structure-property-performance relationships of functional materials by exploring a large library of materials that can also be influenced by wide ranges of compositional, processing, and environmental parameters. The intricate relationship between these factors renders most material science and chemistry experiments a high-dimensional space that is impossible to be exhaustively searched. For maximized productivity and scientific advancement, automation and autonomous experiments~\cite{noack2024methods, pratiush2025scientific, vriza2023self, doerk2023autonomous, yager2023autonomous, waelder2024improved, Szymanski2023-rr, coley2019robotic} 
are thus crucial for selected components of experimentation, instrumentation, and research and development. 
Synchrotrons are particle accelerators that generate extremely intense X-ray beams to enable scientific discoveries by, for example, revealing the internal 3D nano-structure of integrated circuits or human cells as well as capture the dynamics of battery or photovoltaic operation~\cite{willmott2019introduction, barbour2023x, kingan2024structural, holderfield2024concurrent, grunewald2020mapping, holler2019three, lin2017synchrotron}.
User facility instruments that are in high-demand, for example synchrotron beamlines, call for precise control over a suite of hardware and software components and seamless communication between components to allow instrumentation development and easy deployment for broad impact to the general physical science community. 
Advances in AI and machine learning (ML) should be be utilized for efficient and sustainable user facility operations to accelerate material discovery. 
While certain tasks can be fully automated using ML methods, more complex processes require human-instrument collaboration, for example the decision-making process in autonomous experimentation can benefit from both algorithm and human insights.
In software deployment and maintenance, human oversight remains essential. 
Natural language (NL) is the most intuitive way for humans to communicate and can be leveraged to control instrumentation and software to achieve simple yet efficient human-computer interaction (HCI)~\cite{ghaoui2005encyclopedia} and, potentially in the near future, lead to AI-orchestrated holistic scientific discovery.

Rapid development in large language model (LLMs) has led to growing interests in leveraging LLM for accelerating physical science, including in biomedical research~\cite{Swanson2024-va, luo2024intention, meng2024application}, chemistry~\cite{Boiko2023-qu, Bran2023-ms, jablonka2024leveraging, white2023assessment, zheng2023chatgpt}, material design~\cite{okabe2024large, jia2024llmatdesign, chiang2024llamp, reinhart2024large, ghafarollahi2024rapid}, and at scientific user facilities~\cite{tsai2023vision, prince2023opportunities, do2024esac, liu2024synergizing}.
Meanwhile, AI applications in the scientific experimentation pipeline have seen significant advancements with notable progress in the domain of scientific question answering. 
Tools like domain-specific chatbots and Retrieval-Augmented Generation (RAG) systems, such as PaperQA \cite{lala2023paperqa}, have been developed to address the challenge of navigating the growing volume of scientific literature~\cite{yager2023domain, skarlinski2024language}. While these tools are valuable, their scope is typically limited to addressing specific scientific questions, often constituting only an initial step in the scientific experimentation pipeline. For example, systems like PaperQA and its successors demonstrate strong capabilities in literature search and synthesis but do not extend to broader experimental workflows.
Recent advancements extend AI applications beyond question answering to integrated systems for scientific experimentation. The Virtual Lab \cite{Swanson2024-va} employs teams of LLM agents to collaboratively design and validate nanobody binders for SARS-CoV-2 variants, integrating hypothesis generation, computational modeling, and experimental validation. ORGANA \cite{Darvish2024-kv} combines LLMs with robotics to automate diverse chemistry experiments, reducing workload and enhancing efficiency. 
In materials science, \cite{Chen2024-ws} leveraged AI and cloud computing to screen millions of solid electrolyte candidates, identifying promising materials validated experimentally.
Similarly, tools like ChemCrow \cite{Bran2023-ms} and Coscientist \cite{Boiko2023-qu} enhance LLMs with specialized chemistry tools for autonomous experimental planning and execution, bridging computational and experimental domains. 
These systems demonstrate the potential of AI to streamline scientific workflows, integrating multiple stages from hypothesis to experimental validation.
However, many of these systems remain at the prototype or research project stage, often focusing on showcasing specific AI capabilities rather than providing a comprehensive, deployable solution. Real-world scientific workflows require more than the integration of AI models \cite{stoica2024specifications}; they demand robust scaffolding that supports domain-specific customizations, multimodal input interfaces, user-friendly UIs, efficient server-side processing, seamless communication between components, and reliable database management. Without these foundational elements, such systems struggle to transition from proof-of-concept demonstrations to practical tools for experimentalists. 

\begin{figure}[h]
  \centering
  \includegraphics[width=0.5\linewidth]{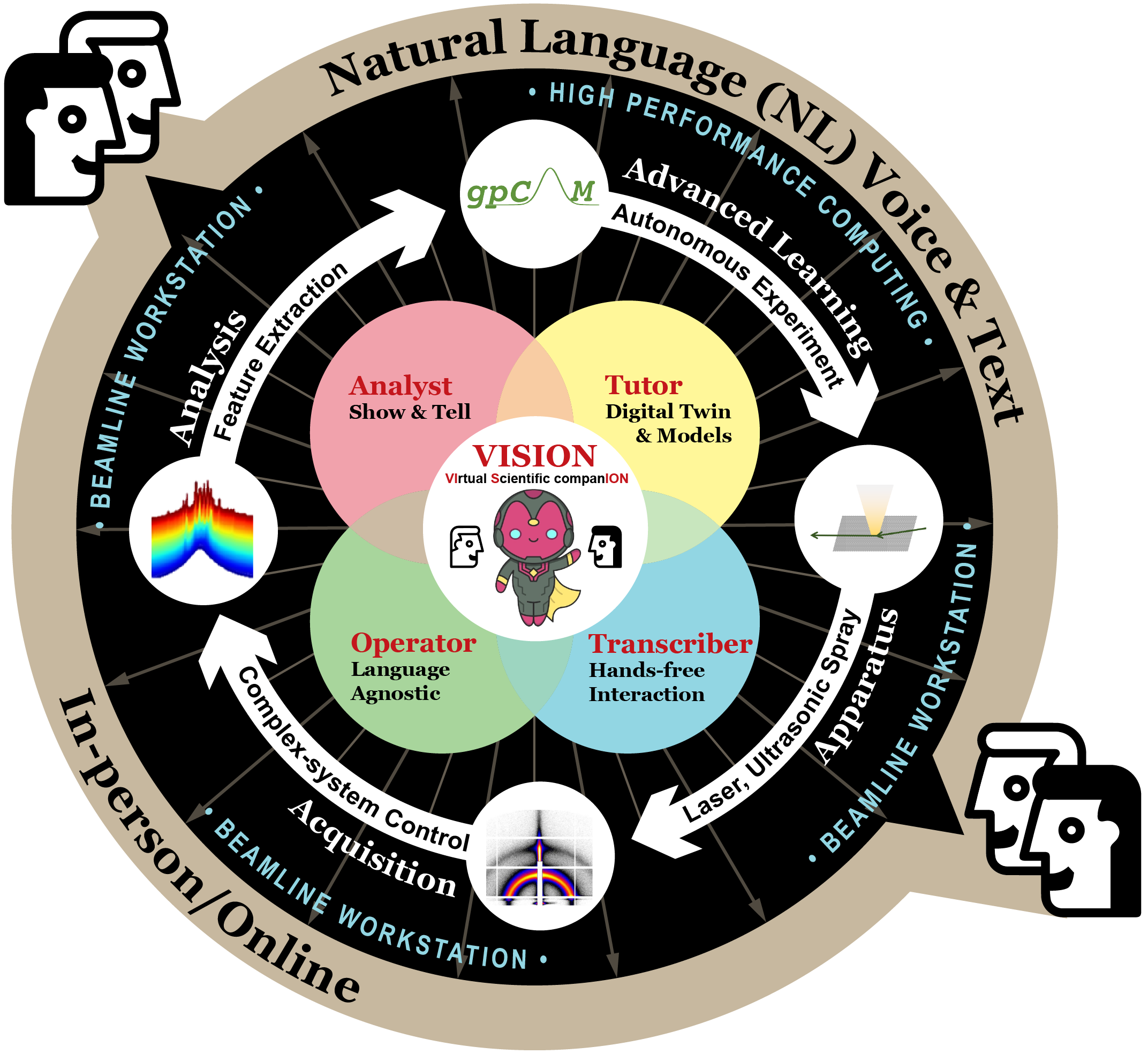}
  \caption{Virtual scientific companion (VISION) aims to lead the
NL-controlled scientific expedition with joint human-AI force
for accelerated scientific discovery at user facilities. 
}
  \label{fig:overview}
\end{figure}

Previously, we have showed the feasibility of utilizing LLMs for data collection at a synchrotron beamline by introducing the prototype of VISION ({\underline{Vi}}rtual
 {\underline{S}}cientific Compan{\underline{ion}})~\cite{tsai2023vision}.
 VISION is a companion whose goal is to assist users on different essential aspects of beamline operation, including acting as a Transcriber allow audio exchange, an Operator to assist with data acquisition, as an Analyst for data analysis, as a Tutor to offer relevant information or guidance during the experiment, as illustrated in Fig.~\ref{fig:overview}.
A domain-specific chatbot was also developed to demonstrate that existing methods and tools can easily be adapted for physical science research~\cite{yager2023domain}.
Here we present an upgraded architecture for VISION to navigate multiple beamline tasks and demonstrate a natural language voice-controlled beamtime.
In the grander scheme, developments in automation and AI-assisted experimentation are fundamental to ultimately building an exocortex---a swarm of AI agents whose intercommunication leads to emergent behavior to extend the cognition and volition of scientists~\cite{kyager2024exocortex}.

The VISION architecture is introduced in Section~\ref{sec:methods}, VISION roles as a Transcriber, Operator, Analyst, and chatbot Tutor are described in Section~\ref{sec:LLMcogs}, whereas a video demonstration\footnote{VISION\_v1 demonstration video available online at \url{https://www.youtube.com/watch?v=NiMLmYVKiQA}} of the voice-controlled beamtime is discussed in Section~\ref{sec:demo}. Section~\ref{sec:conclusion} summarize the work and discuss our future prospects and perspective.


\nopagebreak
\section{Methods}
\label{sec:methods}

Here we present a modular infrastructure for building a practical end-to-end LLM-driven system for scientific experimentation. 
We build a scaffolding on state-of-the-art LLM models to perform domain or instrument-specific tasks.  
The main contributions are as follows:

\begin{itemize}
    \item We address the ambiguity in the literature regarding the definitions of AI agents by introducing the concept of cognitive blocks (cogs) as an abstraction for modular AI functionalities. We clearly differentiate between cogs, assistants, and agents.

    \item We present a modular and scalable architecture composed of an ensemble of cogs tailored for specific tasks. The modularity enables the incorporation of the latest AI models, quick learning of new tasks or diverse workflows, and easy adaptation to other beamlines or instruments, all leading to a scalable system. A relational database was also established to provide long-term monitoring and operational insights.
    
    \item We demonstrate the application of VISION with multi-modal NL-inputs in real-world beamline tasks. The deployment utilized existing beamline light-duty workstation by performing all ML processing on our GPU server HAL ({\underline{H}}PC for {\underline{A}}I and {\underline{L}}LMs). 
    This decoupled beamline GUI and ML processing allows for simple and quick deployment. 

\end{itemize}


\subsection{Terminology: Cognitive Blocks (Cogs), Assistants, and Agents}
\label{sec:term}

In our work, the foundational components of AI systems are defined as \textbf{cognitive blocks (cogs)}. A cog is an individual unit comprising an LLM scaffolded with domain-specific prompts or tools. 
Each cog is designed to perform a specific task or function, such as transcription, classification, or code generation.
Multiple cogs can be put together to form an \textbf{assistant}, where the cogs operate in a pre-defined sequence to accomplish tasks. The sequence of cog execution is deterministic, with the flow of operations specified beforehand based on the user input. 
This structured assembly ensures that each cog contributes its specialized capability while the assistant coordinates their interaction to achieve the desired output. VISION is an AI Assistant that invokes a set of cogs to process natural-language inputs and perform beamline experiments. 

While existing research often refers to systems like VISION as multi-agent systems, we make a distinction between assistant and agent. We reserve the term \textbf{agent} for architectures where multiple cogs interact iteratively to enable adaptive and autonomous behaviors. 
Therefore, VISION's deterministic pre-defined workflow is aligned with the concept of an assistant, while also being a base design for a future agentic AI for instrument control.

\subsection{VISION Architecture}
\label{sec:archi}

\begin{figure}[b]
  \centering
  \includegraphics[width=0.9\linewidth]{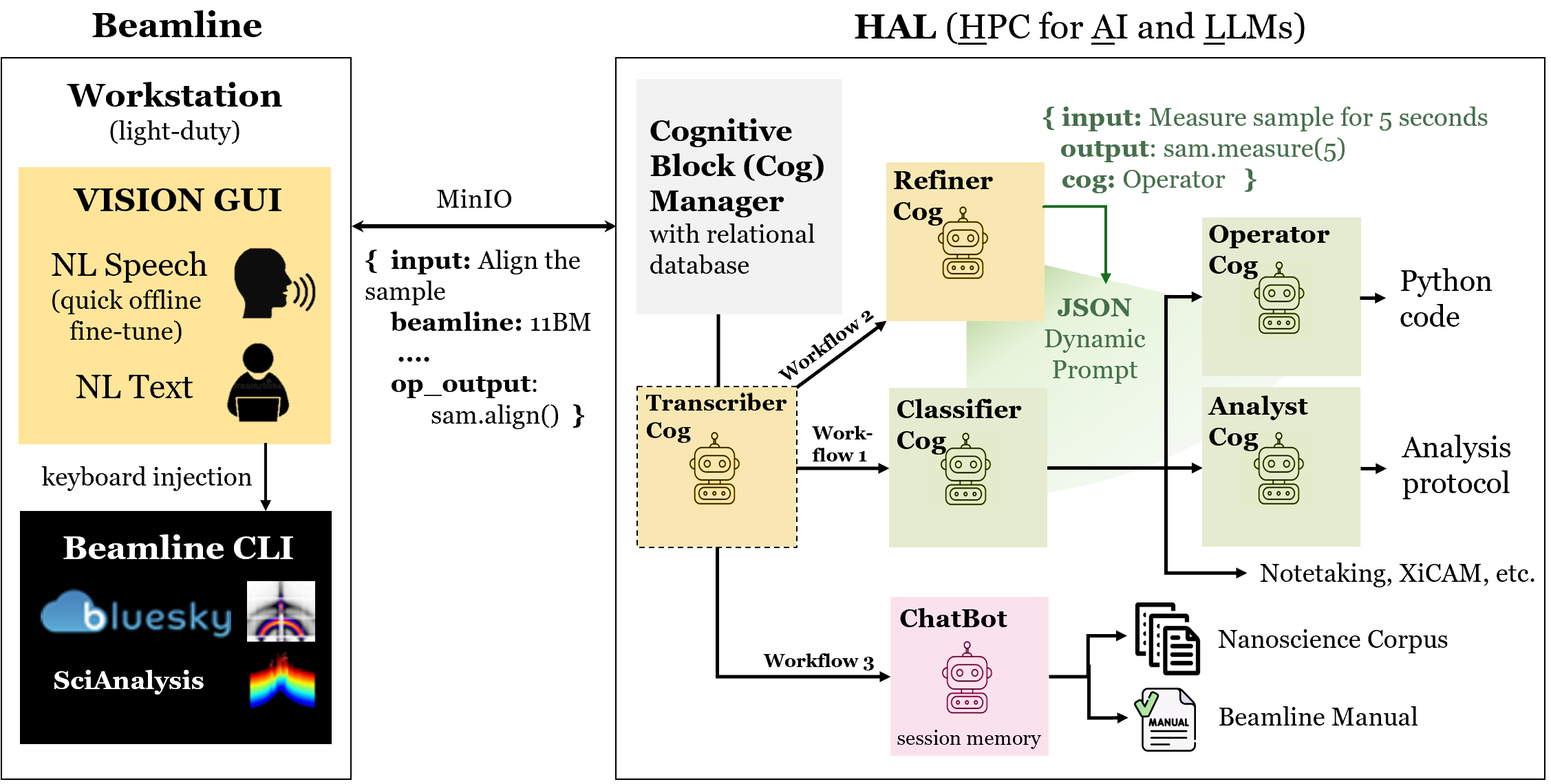}
  \caption{Overview of the VISION architecture, with each cog described in Table~\ref{tab:cog}. }
  \label{fig:architecture}
\end{figure}

VISION is an integrated system designed to process NL text and/or audio input from the beamline user and assign tasks to specialized cogs to produce the desired outputs.
Figure~\ref{fig:architecture} provides an overview of the VISION modular architecture. The key components of the system architecture are as follows:

\begin{itemize}
    \item \textbf{Beamline GUI}: A graphical interface was built using PyQt5, enabling users to interact with VISION via natural language text, speech, or both.
    \item \textbf{Backend Server HAL}: A high-performance server (NVIDIA H100 GPU) was used for fine-tuning and inference of foundation models for the cogs.
    
    \item \textbf{Communication between Beamline and HAL}: This communication is facilitated via MinIO for efficient and secure data exchange. The data is passed in the form of a dictionary with relevant input and output fields.
    
    \item \textbf{Integration with Beamline}: Integration to beamline control is achieved via keystroke injection to communicate with the Bluesky data collection framework~\cite{bluesky}, the SciAnalysis data analysis software~\cite{SciAnalysis_git}, or other existing tools. 
    This provides the advantage of allowing users to use NL-based interaction via VISION while also maintaining the flexibility to use command-line-interface (CLI) as done conventionally. 
    
    \item \textbf{Relational Database}: A database was established (SQLite) to maintain a complete record of VISION's operations, including user inputs, outputs, and actions taken by individual cogs. 
    This comprehensive logging helps monitor system performance, collect user feedback (e.g., whether the result was successfully submitted to the beamline hardware), and identify areas for optimization. Workflows can therefore be refined to reduce response times and enhance overall efficiency.
\end{itemize}

HAL serves as the computational backend for VISION, housing the individual cogs designed to perform specific tasks, shown in Table~\ref{tab:cog}. These cogs are built upon general-purpose models \cite{achiam2023gpt, anthropic2024claude, radford2023robust, yang2024qwen2, hui2024qwen25codertechnicalreport}, which excel at a wide range of tasks but lack inherent knowledge of instrument-specific terminology and workflows. 
To address this, each cog is tailored to include beamline-specific information to effectively execute its designated tasks. The Transcriber cog is fine-tuned to recognize domain- and beamline-specific jargon and terminology, ensuring accurate transcription of audio input. 
For the LLM-based cogs we rely on in-context learning (ICL) \cite{brown2020language} wherein tailored prompts are crafted to provide the necessary beamline context. 
The Classifier cog uses prompts that define command types and specify the corresponding downstream cogs to be invoked. Similarly, the Operator cog relies on a prompt to identify data acquisition and beamline control functions, while the Analyst cog uses a prompt to determine and execute the appropriate analysis protocols. 
A cog prompt can be static (generated before an LLM call) or dynamic (generated at inference).
VISION roles and tasks, including data collection and analysis functions, can vary vastly for each beamline or vary between experiments at the same beamline. 
Since the functions, protocols, and natural language commands can also evolve quickly, we employ a dynamic system prompt building approach. The system prompt for each cog is constructed at inference time from a centralized JSON file, which contains example phrases for potential commands.
During inference, relevant entries from the JSON file are extracted and compiled into the system prompt for the specific cog, as shown in Supplementary Boxes~\ref{lst:examplesjson} and~\ref{lst:finalclassifierprompt}. This approach introduces negligible computational overhead while offering significant flexibility. 
JSON files unique to the experiment or instrument/beamline are placed under corresponding folders. 
Beamline scientists can easily modify or extend system capabilities by updating the JSON file or adding a function using NL through the VISION GUI, minimizing manual effort and ensuring adaptability to new experiments and new instruments in general.
Details on the dynamic prompt generation are provided in Supplementary Section~\ref{SI:dynamic}.

\begin{table}[t] 
    \centering
    \setlength{\tabcolsep}{8pt}
    \caption{VISION Cog Descriptions}
    \label{tab:cog}
    \begin{tabular}{l c c c c}
    \toprule
    \textbf{Cog Type} & 
    \textbf{Cog Name} & 
    \textbf{Model} & 
    \textbf{Open/Closed} &
    \textbf{Provider} \\
    \midrule
    Voice & Transcriber & Whisper-Large-V3 \cite{radford2023robust} & Open & HuggingFace \\
    LLM & Classifier & Qwen2 \cite{yang2024qwen2} & Open & Ollama \\
    LLM & Operator & Qwen2.5-Coder \cite{hui2024qwen25codertechnicalreport} & Open & Ollama \\
    LLM & Analyst & Qwen2 \cite{yang2024qwen2} & Open & Ollama \\
    LLM & Refiner & GPT-4o \cite{openai2023gpt4} & Closed & Azure OpenAI \\
    LLM & Chatbot & GPT-4o \cite{openai2023gpt4} & Closed & Azure OpenAI \\
    \bottomrule
    \end{tabular}
\end{table}

VISION operates as a scientific assistant by invoking cogs in a predefined deterministic order, referred to as workflows here. 
Three types of workflows are supported, each designed to handle specific tasks. 
Depending on the input type, the Cog Manager invokes the appropriate workflow. If the input is audio, the Transcriber cog is invoked first to convert it to text. 
The workflows are labeled in Fig.~\ref{fig:architecture} and described below.

\begin{itemize}
    \item \textbf{Workflow 1: Beamline Commands (Data Acquisition or Analysis)} \\
    Natural language commands are first sent to the Classifier cog, which determines the command type and the next cog to invoke (either Operator or Analyst, or other software tools). For data acquisition commands, the Operator cog generates the corresponding beamline-executable code. For data analysis commands, the Analyst cog retrieves the required analysis protocol. The output from these cogs is sent back to the beamline GUI via the MinIO communication channel for user confirmation. Users can either modify the outputs or confirm the command. Upon confirmation, the generated code is sent to Bluesky for data acquisition or SciAnalysis for data analysis. 

    \item \textbf{Workflow 2: Adding Custom Functions} \\
    This workflow allows users to expand VISION's capabilities by adding new beamline control functions or analysis protocols. Users provide a natural language description of the new function or protocol, which is parsed by the Refiner cog into a structured JSON format. This JSON entry is appended to the centralized JSON file, making the function accessible in upcoming operations through the first workflow.

    \item \textbf{Workflow 3: Chatbot for Nanoscience Queries} \\
    This workflow enables users to ask questions related to nanoscience literature or general topics. The Chatbot cog retrieves relevant answers using its pre-trained knowledge and integration with the nanoscience corpus, ensuring efficient information retrieval.
\end{itemize}

These workflows collectively enable VISION to handle a variety of beamline tasks, demonstrating its flexibility, adaptability, and utility in complex scientific environments. Each workflow is presented to the user as a separate tab on the VISION GUI, shown later in Section~\ref{sec:res}.


\section{Results}
\label{sec:res}

To evaluate the performance of the individual cogs, small evaluation datasets were created for each cog to provide an estimate of their performance. These datasets were not directly included in system prompts, but the evaluation results were used in the development process to refine the system prompts for the specific LLM used.
All evaluations were performed with a sampling temperature of 0. 
Averages were obtained over five runs and, for those that experienced variability, the sample standard deviations of relevant metrics are given (±). 
If there were multiple ground truth answers, the best performing ground truth for each metric is chosen individually when calculating the averages.
Local models are evaluated using Ollama's default quantization (Q4\_0 or Q4\_K\_M) unless otherwise mentioned, with details provided in Supplementary Section~\ref{SI:quantize}.
Throughout this paper, all references to Claude-3.5-Sonnet refer to version 2024-10-22 (via Anthropic API) and all references to GPT-4o refer to version 2024-05-13 (via Azure API).

\subsection{Cog Performance}
\label{sec:LLMcogs}

\subsubsection*{Transcriber Cog}

The {Transcriber cog} of VISION provides speech-to-text functionality built upon OpenAI's Whisper Large-V3 model~\cite{radford2023robust}. While Whisper demonstrates good performance for general English daily phrases, it struggles to recognize beamline-specific jargon critical for beamline operation.
Addressing this limitation, we designed a fine-tuning pipeline for adding domain or beamline-specific terms with fast fine-tuning and without human-recorded audio.
From our studies, we conclude that generic sentence templates can provide sufficient syntactic diversity to encode specialized terms, avoiding the need to craft unique sentences for each term. 
Moreover, synthetic audio generated via Text-to-Speech (TTS) can be used effectively in lieu of human-recorded data, offering a scalable and efficient solution for creating diverse training datasets.
With these two schemes, the fine-tuning pipeline can be easily implemented to offer improved Automatic Speech Recognition (ASR) performance in a parameter-efficient manner~\cite{hu2021lora}, ensuring accurate transcription of specialized terms without significant computational cost.

\begin{figure}[b]
    \centering
    \includegraphics[width=0.7\linewidth]{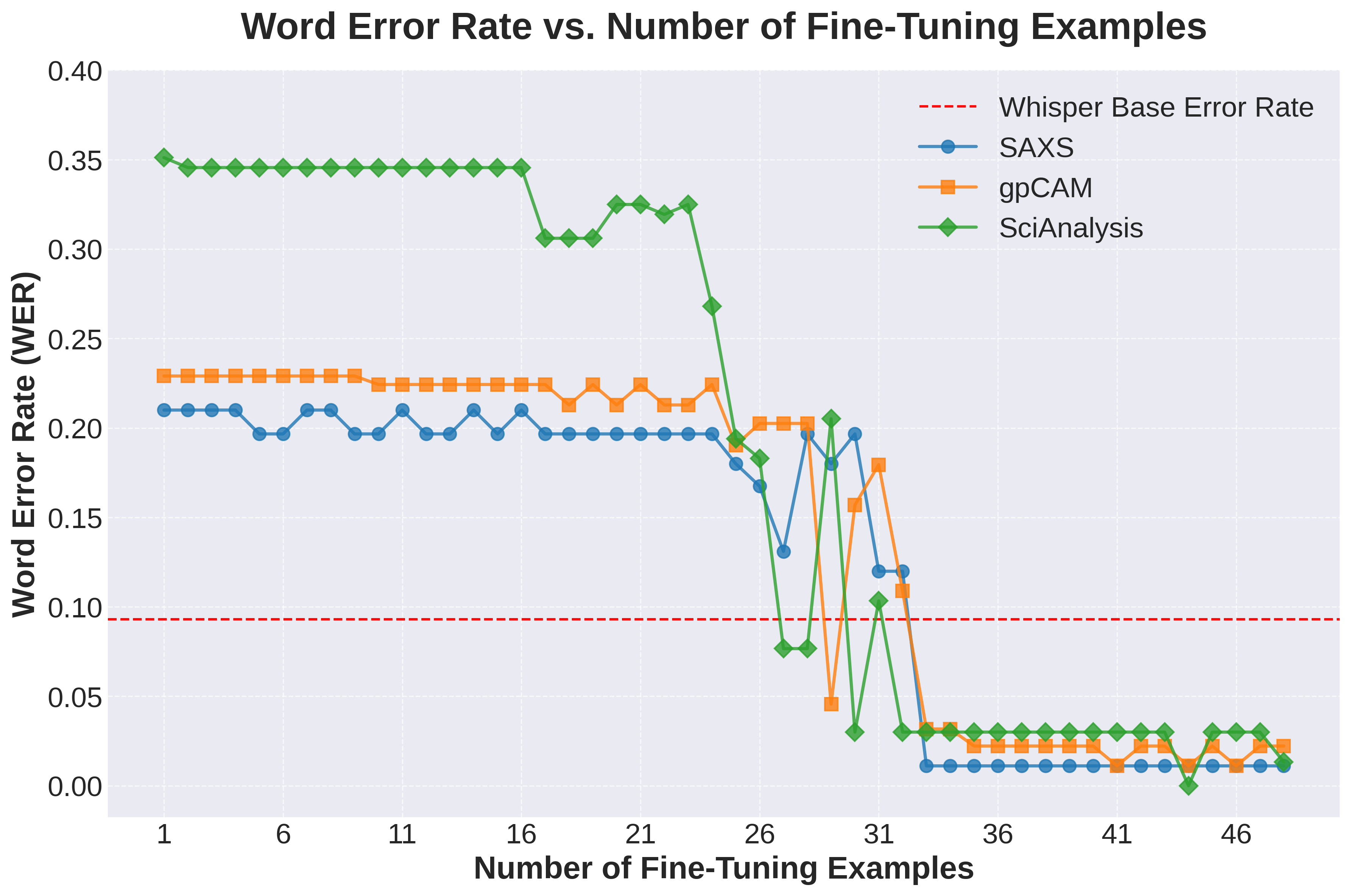}
    \caption{Word Error Rate (WER) performance comparison of fine-tuned Whisper models. The plot shows the WER for beamline specific terms (SAXS, gpCAM, and SciAnalysis) as a function of the number of fine-tuning examples used. The red dashed line represents the baseline WER (0.093) of the Whisper-Large-V3 model.} 
    \label{fig:whisper_wer}
\end{figure}

For each new term, we generated a fine-tuning dataset using around 50 sentence templates and a separate test set using 30 sentence templates (distinct from the fine-tuning dataset). These templates contained a placeholder for the jargon term, which was substituted with the target word (e.g., SAXS, which stands for small angle X-ray scattering). To investigate the minimum number of examples required to teach Whisper an additional word we incrementally increased the number of examples in the fine-tuning dataset for a specific word and evaluated the Word Error Rate (WER) on the test dataset after each step. 
Figure ~\ref{fig:whisper_wer} illustrates our experiment to add three beamline-specific jargon terms: SAXS, gpCAM~\cite{tsuchinoko_git}, and SciAnalysis~\cite{SciAnalysis_git}, wherein the WER on the test set is shown as a function of the number of fine-tuning examples. 
While Whisper already performs well on general English transcription, the initial WER reflects its difficulty in recognizing the specific jargon term. 
The Word Error Rate (WER) decreases as the number of fine-tuning examples increases, with a sharp drop observed after approximately 30 examples. Beyond 40 examples, the WER converges near zero, indicating that Whisper has effectively learned the word. 
The red dashed line represents the baseline WER (0.093) of the Whisper-Large-V3 model for English transcription on the Common Voice 15 dataset \cite{openai_whisper}.
Fine-tuning Whisper for one jargon term on HAL using all 50 sentence templates 
takes around 40 seconds, which remains consistent across words due to the reuse of the same sentence templates for fine-tuning.
This experiment demonstrates that it is feasible to teach individual words to Whisper efficiently. Implementation details are given in Supplementary Section~\ref{SI:whisper}

We have extended this approach to teach Whisper seven jargon terms simultaneously. A combined fine-tuning dataset was created using all the sentence templates for seven words (336 examples in total) and Whisper was fine-tuned in one step, which took about 4 minutes. The final fine-tuned model achieves a WER of zero on the test sets for the individual words, confirming that the model successfully learned all seven terms.

\subsubsection*{Classifier Cog}

Beamline operation includes multiple aspects, as illustrated in Fig.~\ref{fig:overview}, we employ cog classification in VISION to help accomplish specialized tasks that a general cog would likely fail. 
The classifier's task is to interpret user input and determine the next action or cog to call, with five possible classes: \textit{Operator}, \textit{Analyst}, \textit{Note-taker}, and two existing beamline software tools \textit{XiCAM}~\cite{xicam_git} and \textit{gpCAM}~\cite{tsuchinoko_git}. 
The Classifier cog was evaluated using multiple large language models (LLMs) and dynamically generated system prompts. 
The evaluation metrics included F1 score and execution time. The F1 score, which balances precision and recall, was chosen over accuracy due to the imbalanced distribution of data points across categories. 
Execution time represents the duration taken by the classifier to generate a response.

We compared the cog performance by using three types of system prompts:
\begin{itemize}
    \item \textbf{LIST}: The LLM outputs a one-hot encoded vector corresponding to the predicted class.
    \item \textbf{ID}: Each class is assigned a numerical ID (0--4), and the LLM outputs the ID of the predicted class.
    \item \textbf{ONE\_WORD}: The LLM outputs a single word representing the predicted class.
\end{itemize}

Table~\ref{tab:prompt_comparison} summarizes the performance of various LLMs across three prompt types. The ONE\_WORD and ID prompt types consistently achieve the best performance, with high F1 scores and fast execution times. This is likely because these tasks require the LLM to generate very few tokens, in contrast to the LIST prompt type, where the model must perform additional operations to map classes to a one-hot-encoded vector and ensure the vector is in a valid format.
This evaluation highlights the impact of different LLMs and system prompt types on the performance of the Classifier cog. The results provide insight into the trade-offs between accuracy and speed across various configurations.
For example, Athene-V2-Ag achieves top-tier F1 scores for ID and ONE\_WORD prompts but takes significantly longer to process compared to smaller models like Qwen2.5, which balances strong performance with minimal execution time. 
Cog models can easily be swapped out based on the specific accuracy and efficiency needs for a given application.

\begin{table}[b] 
    \centering
    \setlength{\tabcolsep}{8pt}
    \caption{Performance Comparison Across Prompt Types (99 Evaluation Cases)}
    \label{tab:prompt_comparison}
    \begin{threeparttable}
        \begin{tabular}{l c *{6}{c}}
        \toprule
        \multirow{2}{*}{\textbf{Model}} & 
        \multirow{2}{*}{\textbf{Size}} &
        \multicolumn{2}{c}{\textbf{ID}} & 
        \multicolumn{2}{c}{\textbf{ONE\_WORD}} &
        \multicolumn{2}{c}{\textbf{LIST}} \\
        \cmidrule(lr){3-4} \cmidrule(lr){5-6} \cmidrule(lr){7-8}
        &  & \textbf{F1 ($\uparrow$)} & \textbf{Time (s) ($\downarrow$)} 
        & \textbf{F1($\uparrow$)} & \textbf{Time (s) ($\downarrow$)}
        & \textbf{F1($\uparrow$)} & \textbf{Time (s) ($\downarrow$)} \\
        \midrule
        Phi-3.5 & 3.8B & 0.00 & 0.43 ± 0.00 & 0.02 & 1.06 ± 1.44 & 0.00 & 0.40 ± 0.00 \\
        Mistral & 7.25B & 0.67 & 0.32 ± 0.00 & 0.78 & 0.10 ± 0.00 & 0.07 & 3.20 ± 0.55 \\
        Qwen2 & 7.62B & 0.84 & 0.09 ± 0.00 & 0.81 & 0.10 ± 0.00 & 0.32 & 1.63 ± 0.00 \\
        Qwen2.5 & 7.62B & 0.90 & \textbf{0.09 ± 0.00} & 0.90 & 0.09 ± 0.00 & 0.37 & 0.60 ± 0.00 \\
        Mistral-Nemo & 12.2B & 0.86 & 0.14 ± 0.00 & 0.83 & 0.10 ± 0.00 & 0.21 & 0.55 ± 0.00 \\
        Qwen2.5-Coder & 32.8B & 0.94 & 0.14 ± 0.00 & 0.93 & 0.16 ± 0.00 & 0.19 & 4.57 ± 0.15 \\
        Llama3.3 & 70.6B & 0.96 & 0.35 ± 0.00 & 0.95 & 0.40 ± 0.00 & 0.37 & 9.24 ± 0.00 \\
        Athene-V2 & 72.7B & 0.94 & 0.46 ± 0.02 & 0.93 & 0.49 ± 0.01 & 0.49 & 44.48 ± 0.01 \\
        Athene-V2-Ag & 72.7B & 0.95 & 0.20 ± 0.00 & \textbf{0.96} & 0.23 ± 0.00 & 0.53 & 1.95 ± 0.00 \\
        GPT-4o & - & 0.91 & 0.28 ± 0.02 & 0.92 & 0.27 ± 0.03 & 0.91 & 0.52 ± 0.04 \\
        \bottomrule
        \end{tabular}
        \begin{tablenotes}
        \small
        \item \textit{Note:} The F1 score ranges from 0 to 1, with higher values indicating better performance. Among the models tested, only GPT-4o and Phi-3.5 showed variation in F1 scores across multiple runs. The standard deviation (std) for GPT-4o was ±0.01 across all prompt types (ID, ONE\_WORD, and LIST), while Phi-3.5 showed a standard deviation of ±1.44 on only the ONE\_WORD prompt type.
        \end{tablenotes}
    \end{threeparttable}
\end{table}

\begin{figure}[H]
  \centering
  \includegraphics[width=0.9\linewidth]{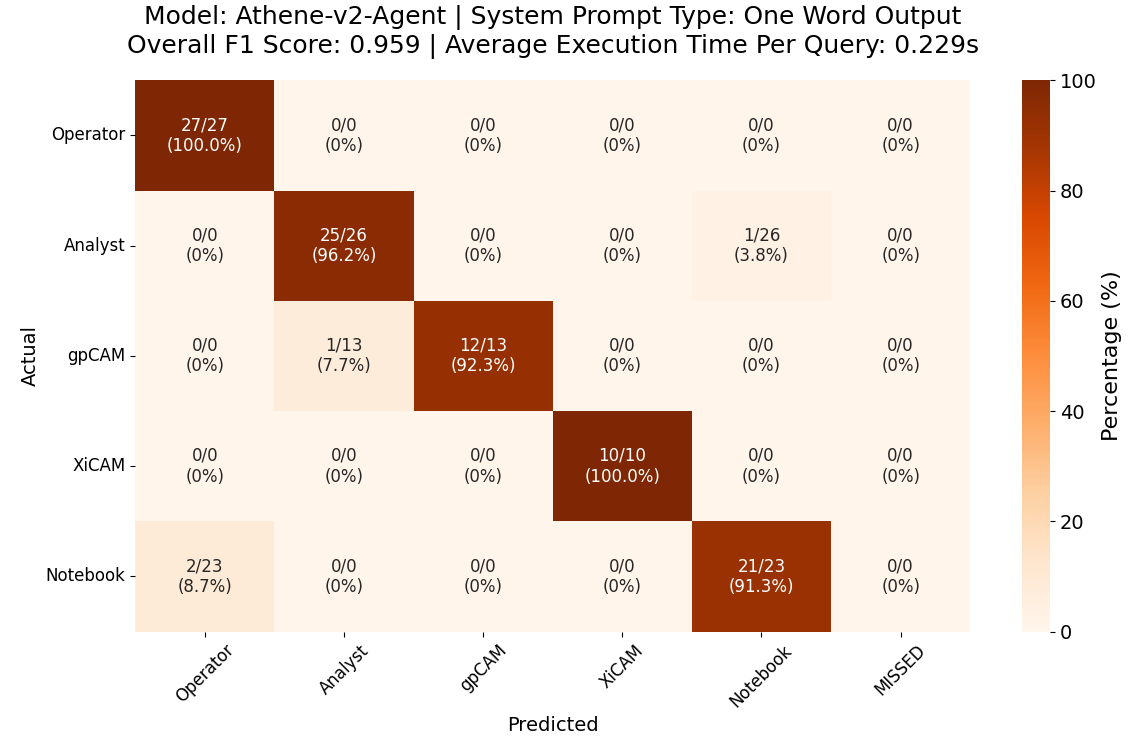}
  \caption{Confusion matrix for Athene-v2-Agent with system prompt type--ONE\_WORD, with failed cases given in Supplementary Table~\ref{tab:failure_cases}.
  }
  \label{fig:athene-v2-agent_cm}
\end{figure}

The confusion matrix for the Classifier cog is given in Fig.~\ref{fig:athene-v2-agent_cm}, showing the performance across five possible classes. Additionally, a \textit{MISSED} class is included to account for instances where the model's predictions do not correspond to any actual class, i.e. possible hallucinated outputs.
The classifier achieves high per-class accuracy, as indicated by the diagonal dominance in the matrix. 
Beamline-related operations are much more restricted compared to general daily conversations and tasks, thus making them easier to classify accurately.
No instances were classified as \textit{MISSED}, demonstrating robust handling of the defined classes. 
The misclassification cases are given in Table~\ref{tab:failure_cases}, which may be due to complex or vague user instruction. 
These ambiguous cases would require clarification from the user or reasoning within the cog to generate the most reasonable output.

Once the classifier identifies the downstream cog or action, VISION can carry out the cog on HAL and/or perform the corresponding action at the beamline. 
For the Analyst cog, user NL-input is categorized as one of the analysis protocols~\cite{SciAnalysis_git} to process the raw 2D scattering data to provide real-time feedback to guide the experiment. Protocols may include circular average or mapping raw data onto the reciprocal space. Here we deployed the most frequently used 8 protocols with results showed in Table~\ref{tab:model-performance-analysis} and details in the Supplementary Section~\ref{SI:ana}.

Existing interactive data visualization tool XiCAM~\cite{pandolfi2018xi} or autonomous experimentation tool gpCAM~\cite{tsuchinoko_git} can also be triggered via VISION given that the Classifier cog can provide correct identification. With a reliable classifier, VISION allows for existing tools and software packages to be set up and used easily using natural language input.

Cog classification for identifying a suitable specialized cog is important as, depending on the tasks, some cogs must be able to reproduce the same outcome, while some cogs can be allowed more freedom. 
The Operator cog has a small constrained set of actions that can be taken and is required to be robust and reliable for instrument control, while the Analyst cog can have a 'higher temperature' and have the liberty to interpret the user intension and provide relevant analysis and visualization.

\subsubsection*{Operator Cog}

The Operator cog is tasked with converting natural language inputs into functional Python code that interfaces with the beamline instruments through customized functions. Here we test on an evaluation dataset that contains sequential control flow examples (simple Python programs with one or two lines of code) and structured control-flow examples (longer programs with conditional statements or loop constructs). The performance was evaluated across multiple LLMs. Note that the system prompt was iteratively improved on this dataset---specifically for \texttt{Qwen2.5-Coder-32B-Instruct} and thus the performance may favor this model. The prompt construction, evaluation, and evaluation metrics are provided in detail in Supplementary Section~\ref{app:opcog}. 

\begin{table}[b] 
    \centering
    \setlength{\tabcolsep}{8pt}
    \caption{Performance for Operator on Sequential Control Flow Examples (120 Evaluation Cases)}
    \label{tab:model-performance-simple}
    \begin{threeparttable}
        \begin{tabular}{l *{4}{c}}
        \toprule
        \textbf{Model} & 
        \textbf{Size} & 
        \textbf{Acc. (\%) ($\uparrow$)} &
        \textbf{Time (s) ($\downarrow$)} &
        \textbf{Norm. Lev. ($\downarrow$)} \\
        \midrule
        Phi-3.5 & 3.8B & 24.17 & 0.6035 ± 0.0012 & 0.3873 \\
        Phi-3.5-fp16 & 3.8B & 20.83 & 0.9400 ± 0.0010 & 0.5351 \\
        Mistral-7B & 7.25B & 31.67 & 0.3808 ± 0.0011 & 0.2788 \\
        Qwen2 & 7.62B & 51.67 & \textbf{0.2771 ± 0.0005} & 0.1683 \\
        Qwen2.5 & 7.62B & 40.83 & 0.2946 ± 0.0010 & 0.2186 \\
        Mistral-NeMo & 12.2B & 29.17 & 0.3744 ± 0.0010 & 0.3107 \\
        Qwen2.5-Coder & 32.8B & 77.50 & 0.7288 ± 0.0021 & 0.1012 \\
        Llama-3.3 & 70.6B & 66.67 & 1.3401 ± 0.0018 & 0.1208 \\
        Athene-v2 & 72.7B & 78.33 & 1.2465 ± 0.0026 & 0.0881 \\
        Athene-v2-Agent & 72.7B & 77.50 & 1.2299 ± 0.0002 & 0.0731 \\
        Claude-3.5-Sonnet & - & \textbf{83.33} & 1.5378 ± 0.1592 & \textbf{0.0555} \\
        GPT-4o & - & 82.33 & 0.4259 ± 0.0321 & 0.0574\\
        \bottomrule
        \end{tabular}
        \begin{tablenotes}
      \small
      \item \textit{Note:} Size (B) represents model parameters in billions. Normalized Levenshtein Distance (Norm. Lev.) ranges from 0 to 1, lower is better. GPT-4o showed variation in accuracy (std: ±0.9129) and Norm. Lev. (std: ±0.0028). Averaged over 5 runs. See \ref{sup:op_metrics} for metrics definition.
    \end{tablenotes}
    \end{threeparttable}
\end{table}

Table \ref{tab:model-performance-simple} shows the results on the sequential control flow evaluation dataset. 
These types of problems generally contain one or two NL short sentences that the model needs to convert to Python code. It does this by using the function descriptions in its system prompt and filling in the correct variables.
The results show several trends: larger models (>30B parameters) generally achieve higher accuracy (>75\%) and Normalized Levenshtein Distance (a metric between 0 and 1 measuring the similarity between sequences, where 0 indicates identical sequences), with \texttt{Claude-3.5-Sonnet} leading at 83.3\%. A speed and size trade-off is evident: smaller models like \texttt{Qwen2} (7.62B) are fastest (0.28s), while larger models are slower but more accurate. \texttt{Qwen2.5-coder} seems to punch above its weight for its size, however this might be because its evaluation results were used to improve the system prompt. 
The Phi-3.5 models were significantly slower, this is because they produce a lot of tokens trying to explain themselves (sometimes in comments---which are stripped out, other times not in comments---which results in incorrect code), which affects the inference time and possibly accuracy too. 

Additionally, for data acquisition requests which require structured control flow with loop constructs and conditional statements, some extra examples have been written and tested. Table~\ref{tab:model-performance-complex} shows results with 
CodeBLEU \cite{ren2020codebleumethodautomaticevaluation} as the primary metric and 
Box~\ref{lst:structured-control-flow} shows an example of the output of Claude-3.5-Sonnet on this portion of the dataset. 
The ground truth solutions were partially constructed from outputs of \texttt{Qwen2.5-Coder} and \texttt{Claude-3.5-Sonnet}, which may bias the metrics in their favor. 
The results show that \texttt{Qwen2.5-Coder} performs best across metrics for structured tasks, achieving two exact matches and the highest CodeBLEU score of 0.79. Smaller models (<10B parameters) struggle significantly with these complex tasks, though \texttt{Qwen2.5} shows surprisingly good performance for its size. 
Inference times are notably higher for structured tasks than for sequential tasks, with larger models taking multiple seconds per example.
System mocking capabilities for evaluating functional equivalence are under development to provide an accurate assessment on the code execution outcome as opposed to code similarity metrics.

\begin{lstlisting}[style=pythonstyle, label={lst:structured-control-flow}, caption={Example of a structured control flow with Python code generated by Claude-3-5-Sonnet (2024-10-22) in response to the prompt ``Measure the sample for 1s (every 10s), do this for 1 min''.}]
import time

start_time = time.time()
end_time = start_time + 60

while time.time() < end_time:
    loop_start = time.time()
    sam.measure(1)
    elapsed = time.time() - loop_start
    if elapsed < 10:
        time.sleep(10 - elapsed)
\end{lstlisting}


\subsection{Beamline Demo}
\label{sec:demo}

We have deployed VISION at the 11-BM Complex Materials Scattering (CMS) beamline at the National Synchrotron Light Source II (NSLS-II) at the Brookhaven National Laboratory. We used VISION to measure a liquid crystal polymer thin film and observe changes in its crystalline phases as temperature varied. Grazing-incidence wide angle X-ray scattering (GIWAXS) characterizations were performed at the CMS beamline. Thin film samples were measured at incident angle 0.14 deg with a $200 \, \mu \mathrm{m} \times 50 \, \mu \mathrm{m}$ (H$\times$V) beam at wavelength $\lambda=0.9184$~\AA (13.5 keV). The 2D scattering patterns were obtained using Dectris detector Pilatus800k positioned 0.26~m downstream of the samples. Data analysis was carried out using beamline-developed software SciAnalysis~\cite{SciAnalysis_git}.
As described in Fig.~\ref{fig:architecture}, the VISION GUI was launched on the beamline workstation, the computing work and LLM operations were performed or initiated on HAL. 
Without interfering with the interactive IPython Bluesky~\cite{bluesky} terminal, we used keyboard injection for VISION to interact with Bluesky for controlling beamline instruments to maintain the ability to use conventional CLI.

\begin{figure}[h]
  \centering
  \includegraphics[width=0.95\linewidth]{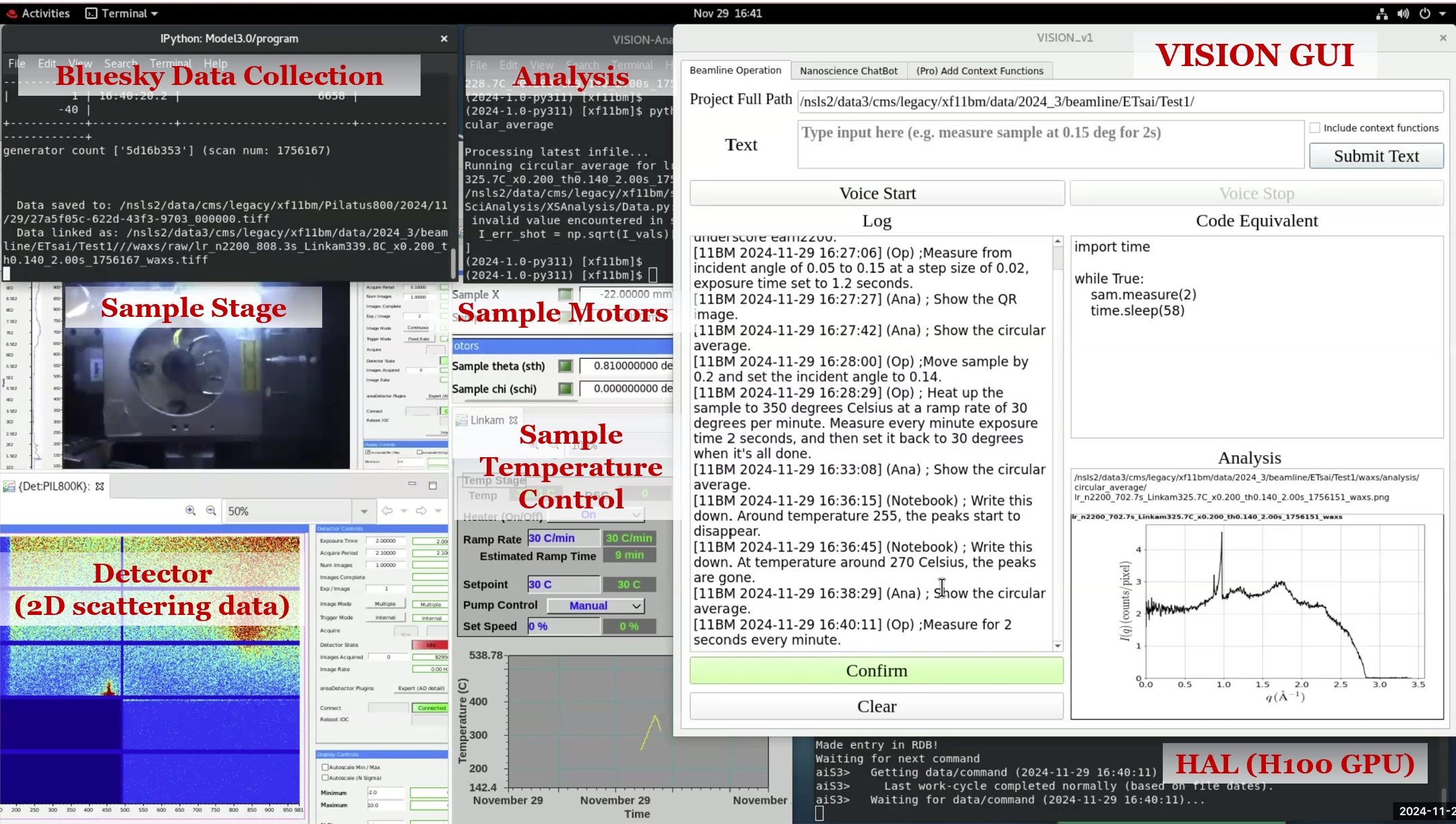}
  \caption{VISION deployment at NSLS-II 11-BM CMS: GUI was launched at the beamline workstation, with backend processing performed on HAL. LLM-based cog results are displayed to prompt user confirmation, followed by execution on Bluesky or other software.
  }
  \label{fig:demo_1}
\end{figure}

We demonstrated and recorded the first voice-controlled beamtime (experiment time at a beamline) to show the NL-controlled data acquisition code generation, basic data analysis, adding new function to a cog, and a domain-specific chatbot. As shown in Fig.~\ref{fig:demo_1}, the GUI allows the user to type/speak natural language, or both. The identified cog and task are displayed under the log box with a timestamp---once the user confirms an action, the code equivalent will be sent to the corresponding terminal for execution. 
The video clip\footnote{VISION\_v1 demonstration video available online at \url{https://www.youtube.com/watch?v=NiMLmYVKiQA}} shows in real-time that the user can speak in natural language to move the sample motor, trigger the detector, control the sample stage temperature, and perform basic data analysis and visualization. The user can also use VISION as a voice recorder to take notes during the experiment, e.g. \textit{around $255^{\circ}\mathrm{C}$ we observed phase transition}, which is saved to a spreadsheet as a CSV file for convenient browsing. 

While the data collection and analysis workflow is given in the first tab of the GUI, 
Fig.~\ref{fig:demo_2} in Supplementary Section~\ref{SI:demo} shows the workflow for the chatbot on the second tab and the add-custom-function workflow on the third tab.
For VISION to be deployed efficiently at a new beamline and for LLM cogs to learn about new beamline functions, the add-custom-function tab allows new functions to be added via text and/or audio, which is converted to the JSON format for dynamic prompting. 
Users can also leverage the chatbot tab to ask beamline-specific, domain-specific, or general questions. 
The nanoscience chatbot provides responses based on query type by classifying queries as either scientific or generic, then it decides if a short scientific answer or a thorough answer based on scientific references~\cite{lala2023paperqa} is suitable. This multi-step determination process involves some stochasticity, which mimics the intrinsic randomness in real conversations. Details on the chatbot workflow is provided in Supplementary Section~\ref{SI:chatbot}. Cog prompts used are listed in Supplementary Section~\ref{SI:prompts}.

\section{Discussion and Conclusion}
\label{sec:conclusion}

To go beyond basic sequential and structured control flows (e.g. \texttt{for} or \texttt{while} loops), code verification under a mock environment for the beamline data collection framework Bluesky~\cite{bluesky} has to be developed. A mock environment will additionally allow VISION to write complex Python scripts and to improve them with reflection. The Operator cog's capabilities can be significantly expanded once the evaluation can be done reliably and fully automated with true functional equivalence testing. Moreover, the code equivalent box allows users to edit the code, therefore all modifications can be logged for further prompt development and performance improvement. 
Realistically, NL-inputs from users, especially new users, are at times ambiguous or incomplete and thus developments are undergoing for VISION to collect essential but missing information from users via dialog. It is important for both the Operator and Analyst cogs to have the capability of interacting with users iteratively to ensure user's demands are met. ASR model should continue to be improved to allow fast and smooth communication across all domain-specific tasks.
User NL-inputs can also be broad in scope and thus require VISION to ask followup questions or breakdown a goal into sequences of tasks.
In order to provide proper guidance or engage in true scientific discussions, complete sets of literature, instrument capabilities and documentation, text-formatted codebase, and correlated multi-model data should also be included. 
Continuous development of the Classifier cog is also crucial to capture multiple/all beamline hardware and software capabilities. Accurate classification also allows cogs to have different 'temperature' depending on the nature of their specialized tasks. 
Interface with existing beamline tools is currently done with keyboard injection, which gives minimal friction to introducing VISION since it does not replace the conventional CLI for interactive IPython workflow. However, this implementation is not robust and an enhanced integration is needed by, for example, leveraging the Bluesky Queue Server for a more structured workflow or building a combined CLI and VISION GUI with a shared namespace. 


In this work, VISION has been tailored to address the specific needs of particular beamline operations, specifically standard X-ray scattering experiments for characterizing nanomaterials and soft matters.
Its modular design results in a scalable system through the easy integration of the latest AI models, smooth adaptation to other beamlines or instruments, and quick learning of new tasks or workflows.
This modularity and scalability support VISION to be readily adapted and expanded to support a broader range of beamline applications or other complex instrumentation, for example multi-model in-situ characterization, spectroscopy or imaging at synchrotron beamlines or electron microscopy instrumentation. 

%
The development of VISION marks a significant step toward realizing the broader scheme of a scientific exocortex ~\cite{kyager2024exocortex}. The exocortex presents an idea of an integrated network of AI systems designed to augment researchers' cognitive and operational capabilities. Within this paradigm, VISION is an assistant that enables NL-based beamline operation and thus also encourages scientists to focus on higher-level scientific inquiries.
The benefits of AI/ML advances towards physical science can be viewed as twofold: one is to assist in automation for repetitive tasks or standard operations, intrinsically doing what state-of-the-art instrument or human scientists can already do but much faster and more sustainable; the other is to augment human intelligence with AI by building a scientific exocortex, intrinsically doing what is not (yet) possible.
NL-based VISION is an effort towards the first mission, while also laying foundation for the second. 
To enable transformative science through AI, advances in LLMs can be the solution to existing gaps in scientific infrastructure and research culture by connecting existing but disperse software and hardware capabilities, connecting feasibility to productivity with painless deployment, connecting humans to instrumentation and resources with intuitive and practical interfaces, connecting multi-disciplinary scientists for collective inspiration and collaborative efforts, connecting the general public to scientific thinking by lowering the learning barrier, and most of all, connecting us to a future with AI-augmented humans for scientific discovery and breakthroughs.

\subsection*{Conclusion}

VISION marks a milestone as the first voice-controlled beamline system by running LLM-based operations on a light-duty beamline workstation with low latency, pioneering a novel approach for beamline operation. 
The modular and scalable architecture enables VISION to navigate multiple roles and tasks seamlessly while also allowing continuous development of new instrument capabilities and requiring minimal effort to adapt VISION to new instrumentation.  
We envision VISION to synergistically partner humans with computer controls for augmented human-AI-facility performance at synchrotron beamlines and beyond, leading the way to a new era where NL-based communication will be the primary interface for scientific experimentation while also paving the roads towards AI-augmented discovery via a scientific exocortex.


\section*{Acknowledgments}
The work was supported by a DOE Early Career Research Program.
This research also used beamline 11BM (CMS) of the National Synchrotron Light Source II and utilized the X-ray scattering partner user program at the Center for Functional Nanomaterials (CFN), both of which are U.S. Department of Energy (DOE) Office of Science User Facilities operated for the DOE Office of Science by Brookhaven National Laboratory under Contract No. DE-SC0012704. 
We thank beamline scientist Dr. Ruipeng Li for consulting the project and his support at the beamline and Dr. Lee Richter for providing the polymer sample for the beamtime.

\bibliographystyle{unsrt}  
\bibliography{NLP, xray}



\beginsupplement

\section{Supplementary Information}
\label{sec:SI}

Method details are described as follows.
Section~\ref{SI:whisper} provides fine-tuning of Whisper, 
Section~\ref{SI:LLMcogs} gives implementation details of the Classifier cog, the Operator cog, and the Analyst cog.
The nanoscience chatbot workflow is described in Section~\ref{SI:chatbot} and snapshots of the beamline demo are shown in Section~\ref{SI:demo}. The prompts used for the cogs are described in Section~\ref{SI:prompts}.

\subsection{Fine-Tuning Whisper for Domain-Specific Terminology}
\label{SI:whisper}

\begin{figure}[H]
  \centering
  \includegraphics[width=0.9\linewidth]{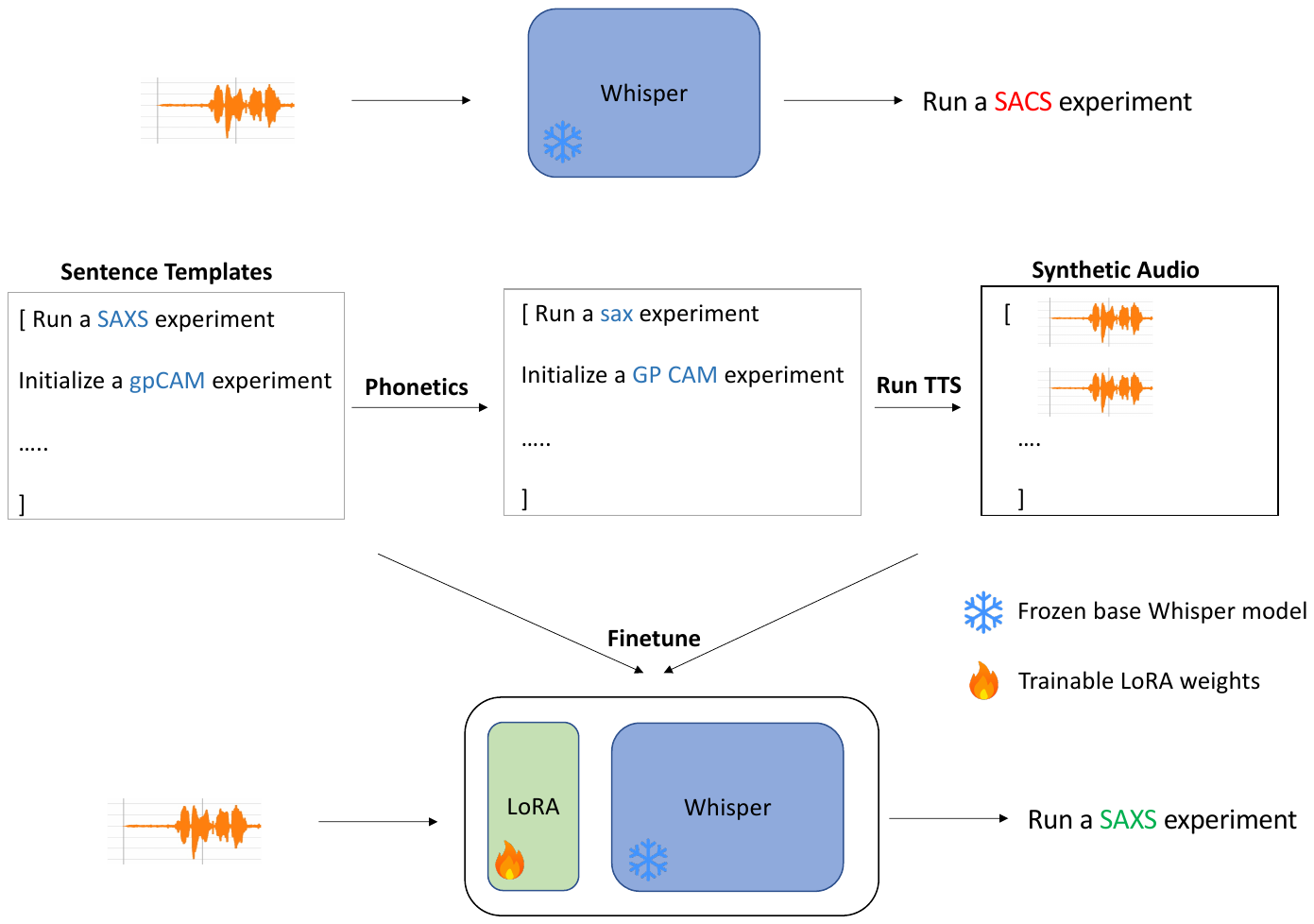
  }
  \caption{Voice fine-tuning pipeline}
  \label{fig:voice_finetuning}
\end{figure}

Previous works have tackled the broader challenges of teaching Whisper low-resource languages \cite{song2024lora, xu2024towards, polat2024implementation} and adapting it to new domains using synthetic data \cite{vasquez2023whisper}. However, the more specific question of \textit{what is required to teach Whisper additional terms or jargon} is relatively underexplored. 
In this work, we explore this question by combining synthetic audio generation and Low-Rank Adaptation (LoRA) \cite{hu2021lora} to determine the minimum number of data points necessary to encode domain-specific terms effectively. 
To enable quick and easy addition of new terms, we propose the following two schemes, which are proven to be useful in our case to add beamline-specific terms. 

\begin{itemize}
    \item \textbf{Sentence Templates for Encoding Specialized Terms}: 
    We explored the importance of sentence context in the fine-tuning dataset. General sentence templates unrelated to the target word's typical context were used in the fine-tuning dataset, while the test dataset consisted of distinct sentence structures. This setup allowed us to evaluate whether Whisper focuses on learning the word itself or depends on contextual clues.
    
    \item \textbf{Synthetic Audio for Scalability}: 
    We examined the feasibility of using synthetic audio for fine-tuning. Synthetic audio samples of the target word were included in the fine-tuning dataset, and the model's performance on real-world voices in the test dataset was analyzed to determine its ability to generalize across audio sources.
\end{itemize}

The fine-tuning pipeline consisted of the following steps:

\begin{enumerate}
    \item \textbf{Generic Sentence Templates}: Approximately 50 sentence templates were crafted to provide diverse syntactic contexts for the target jargon. These templates were adaptable to any specialized term, with examples like \textit{"The term [X] is used in..."} or \textit{"[X] plays a critical role in..."}, allowing seamless substitution to create a varied training corpus.
    
    \item \textbf{Synthetic Audio Generation}: The modified sentences were fed into Google Translate’s Text-to-Speech (TTS) API to generate synthetic audio clips. By spelling jargon terms phonetically, the generated audio closely mimicked the pronunciation used by domain specialists, ensuring alignment with real-world usage.
    
    \item \textbf{Dataset Assembly}: 
    For each new domain-specific term or word, approximately 50 text-audio pairs were used for fine-tuning and 30 text-audio pairs for testing. For a vocabulary of $n$ new words 
    and $k$ sentences per word, the dataset contained $n \times k$ text-audio pairs. This efficient design balanced dataset size and diversity. Training and testing datasets use different sentence templates.
    
    \item \textbf{Fine-Tuning via LoRA}: The assembled dataset was used to fine-tune Whisper using LoRA, a parameter-efficient approach that integrates new knowledge without requiring full model retraining. LoRA’s ability to minimize computational overhead while effectively adapting models has been demonstrated in prior work \cite{song2024lora, xu2024towards, polat2024implementation}.
\end{enumerate}

The fine-tuning training dataset consisted of audio-text pairs designed to teach the Whisper model beamline-specific terms, while the testing dataset was kept separate to evaluate the model's ability to generalize to unknown examples.

Our exploration highlights the feasibility of encoding specialized terminology into Whisper with minimal data requirements. By presenting this methodology, we aim to contribute to the broader understanding of fine-tuning strategies for domain-specific terminology in speech-to-text systems.

\subsection{LLM Cogs}
\label{SI:LLMcogs}

\subsubsection{Dynamic System Prompts with JSON}
\label{SI:dynamic}

The system prompts for the LLM cogs are constructed dynamically at runtime. Each system prompt consists of two components: a base prompt (Box \ref{lst:classifierprompt}) and a set of examples (Box \ref{lst:examplesjson}). 

The base prompt contains general instructions specifying the behavior of the LLM cog. These instructions ensure consistent processing of user inputs and responses. The examples are task-specific input-output pairs stored in a JSON file. Each cog accesses only the relevant examples corresponding to its functionality.

At runtime, the system prompt is assembled by appending the relevant examples to the base prompt (Box \ref{lst:finalclassifierprompt}). The base prompts are stored in a centralized file, while examples for all cogs are managed in a separate JSON file. This separation simplifies updates: users can modify general behavior by editing the base prompts file or add new examples by updating the JSON file.

This approach provides a structured and modular design, allowing the system to adapt to changes in user requirements or domain-specific tasks with minimal overhead.

\begin{lstlisting}[language=Python, caption={Classifier Base Prompt}, label=lst:classifierprompt]
classifier_cog = (

'''
You are a classifier agent designed to analyze user prompts. Process them according to the following instructions:
    
1. Determine the Command Type:

   - Operator (Op):
     - Any task that involves hardware control

   - Analyst (Ana):
     - Data analysis tasks

   - Notebook:
     - Logging tasks, or writing tasks, or general observations

   - gpCAM (gpcam):
     - Only predict gpcam if you see it in the human input

   - xiCAM (xicam): 
     - Only predict xicam if you see it in the human input

2. Analyze the user prompt and determine whether the command is an Op, Ana, Notebook, gpcam, or xicam.

3. Strictly provide the output only in the following format:

   - Output only one word indicating the class:
     - Op
     - Ana
     - Notebook
     - gpcam
     - xicam
    
Use the following examples to learn about how to generate your outputs:

'''
)
\end{lstlisting}

\begin{lstlisting}[language=json, caption={Examples JSON}, label=lst:examplesjson]
[
    {
        "example_inputs": ["Measure the sample for 5 seconds and increase the temperature by 10 degrees."],
        "output": "",
        "cog": "Op",
        "default": true
    },
    {
        "example_inputs": ["Analyze the diffraction patterns and perform peak fitting."],
        "output": "",
        "cog": "Ana",
        "default": true
    },
    {
        "example_inputs": ["Set the humidity to 60% and start the beamline measurement."],
        "output": "",
        "cog": "Op",
        "default": true
    },
    {
        "example_inputs": ["Convert the detector images to reciprocal space q images and perform data reduction."],
        "output": "",
        "cog": "Ana",
        "default": true
    },
    {
        "example_inputs": ["Record that we observed unexpected peaks at high temperatures."],
        "output": "",
        "cog": "Notebook",
        "default": true
    },
    {
        "example_inputs": ["Note: The sample alignment was off by 2 degrees."],
        "output": "",
        "cog": "Notebook",
        "default": true
    }

]
\end{lstlisting}

\begin{lstlisting}[language={}, caption={Example of a Classifier prompt with dynamic prompt generation}, label=lst:finalclassifierprompt]
System: 
You are a classifier agent designed to analyze user prompts. Process them according to the following instructions:
    
1. Determine the Command Type:

   - Operator (Op):
     - Any task that involves hardware control

   - Analyst (Ana):
     - Data analysis tasks

   - Notebook:
     - Logging tasks, or writing tasks, or general observations

   - gpCAM (gpcam):
     - Only predict gpcam if you see it in the human input

   - xiCAM (xicam): 
     - Only predict xicam if you see it in the human input

2. Analyze the user prompt and determine whether the command is an Op, Ana, Notebook, gpcam, or xicam.

3. Strictly provide the output only in the following format:

   - Output only one word indicating the class:
     - Op
     - Ana
     - Notebook
     - gpcam
     - xicam

4. Always output one word corresponding to the identified class.
    
Use the following examples to learn about how to generate your outputs:

Examples:
Example 1:
User Prompt: Measure the sample for 5 seconds and increase the temperature by 10 degrees.
Your Output: Op

Example 2:
User Prompt: Analyze the diffraction patterns and perform peak fitting.
Your Output: Ana

Example 3:
User Prompt: Set the humidity to 60% and start the beamline measurement.
Your Output: Op

Example 4:
User Prompt: Convert the detector images to reciprocal space q images and perform data reduction.
Your Output: Ana

Example 5:
User Prompt: Record that we observed unexpected peaks at high temperatures.
Your Output: Notebook

Example 6:
User Prompt: Note: The sample alignment was off by 2 degrees.
Your Output: Notebook
\end{lstlisting}


\subsubsection{Quantizations per Model}
\label{SI:quantize}
The majority of models were deployed using default quantization settings through Ollama\footnote{\url{https://ollama.com/}}. Table \ref{tab:quantizations} specifies the quantization method used for each model.

\begin{table}[H]
    \centering
    \setlength{\tabcolsep}{8pt}
    \caption{Model Quantization Specifications}
    \label{tab:quantizations}
    \begin{threeparttable}
        \begin{tabular}{l c}
        \toprule
        \textbf{Model} & \textbf{Quantization} \\
        \midrule
        Phi-3.5 & Q4\_0 \\
        Phi-3.5-fp16 & FP16 \\
        Mistral-7B & Q4\_0 \\
        Qwen2 & Q4\_0 \\
        Qwen2.5 & Q4\_K\_M \\
        Mistral-NeMo & Q4\_0 \\
        Qwen2.5-Coder & Q4\_K\_M \\
        Llama-3.3 & Q4\_K\_M \\
        Athene-v2 & Q4\_K\_M \\
        Athene-v2-Agent\footnotemark & Q4\_K\_M \\
        Claude-3.5-Sonnet & N/A \\
        GPT-4o & N/A \\
        \bottomrule
        \end{tabular}
    \end{threeparttable}
\end{table}
\footnotetext{Available at \url{hf.co/lmstudio-community/Athene-V2-Agent-GGUF}}


\subsubsection{Classifier Cog}

The Classifier cog is responsible for determining the type of task a user intends to perform based on their input, which can be provided as either pure text and/or transcriptions generated by the Transcriber cog. This classification helps ensure that the user commands are directed to the appropriate cog within VISION. Specifically, it determines whether the input requires generating relevant code through the Operator or Analyst cogs, initiating experiments with XiCAM or gpCAM, or performing note-taking functions. To achieve this, the Classifier cog leverages Qwen2 \cite{yang2024qwen2} as the underlying LLM. The detailed system prompt used for this cog is given in \ref{prompts:classifier}. 

Table \ref{tab:failure_cases} shows the failure cases for Athene-v2-Agent which achieves the highest F1 score (Table \ref{tab:prompt_comparison}) of the evaluated models. 



\begin{table}[H]
\centering
\begin{tabular}{|p{6cm}|l|l|}
\hline
\textbf{Input Command} & \textbf{Expected Output} & \textbf{Generated Output} \\ \hline
The signal is very weak, we can't really see the peaks. & XiCAM & Operator \\ \hline
Make the thumbnail images use the vge\_hdr color map. & Analyst & XiCAM \\ \hline
Change the objective function to use Shannon entropy. & gpCAM & Analyst \\ \hline
Loading sample BZ02. Sample is oriented vertically. & XiCAM & Operator \\ \hline
\end{tabular}
\caption{Failure cases observed for Athene-v2-Agent as the classifier}
\label{tab:failure_cases}
\end{table}

\subsubsection{Analyst}
\label{SI:ana}

The Analyst cog leverages SciAnalysis to analyze experimental data collected through the Operator cog's interactions with beamline instruments, enabling visualization through plots and quantitative analysis through e.g. circular integration and line cuts.
Table \ref{tab:model-performance-analysis} shows results for the Analyst cog evaluation across 37 evaluation cases. While larger models generally perform better, with \texttt{Athene-v2-Agent} and \texttt{Qwen2.5-Coder} leading in accuracy and edit distance metrics. 
Mid-sized models like \texttt{Qwen2.5} show competitive performance, and inference times remain practical across all models except for \texttt{Claude-3.5-Sonnet}.

\begin{table}[htbp]
    \centering
    \setlength{\tabcolsep}{8pt}
    \caption{Performance for Analysis Tasks on 37 Evaluation Cases}
    \label{tab:model-performance-analysis}
    \begin{threeparttable}
        \begin{tabular}{l *{5}{c}}
        \toprule
        \textbf{Model} & 
        \textbf{Size} & 
        \textbf{Acc. (\%) ($\uparrow$)} &
        \textbf{Time (s) ($\downarrow$)} &
        \textbf{Lev. ($\downarrow$)} &
        \textbf{Norm. Lev. ($\downarrow$)} \\
        \midrule
        Phi-3.5 & 3.8B & 21.62 & 0.3714 ± 0.0017 & 194.43 & 0.6950 \\
        Phi-3.5-fp16 & 3.8B & 32.43 & 0.3947 ± 0.0017 & 160.46 & 0.6066 \\
        Mistral-7B & 7.25B & 48.65 & \textbf{0.0711} ± 0.0005 & 4.16 & 0.3009 \\
        Qwen2 & 7.62B & 62.16 & 0.0832 ± 0.0014 & 3.11 & 0.2215 \\
        Qwen2.5 & 7.62B & 70.27 & 0.0823 ± 0.0086 & 2.24 & 0.1875 \\
        Mistral-NeMo & 12.2B & 45.95 & 0.1195 ± 0.0099 & 6.00 & 0.2910 \\
        Qwen2.5-Coder & 32.8B & 78.38 & 0.1532 ± 0.0028 & 1.54 & 0.1362 \\
        Llama-3.3 & 70.6B & 64.86 & 0.2220 ± 0.0022 & 2.78 & 0.1989 \\
        Athene-v2 & 72.7B & 75.68 & 0.2380 ± 0.0068 & 2.03 & 0.1294 \\
        Athene-v2-Agent & 72.7B & \textbf{81.08} & 0.2530 ± 0.0243 & \textbf{1.35} & \textbf{0.1058} \\
        Claude-3.5-Sonnet & - & 72.97 & 1.2458 ± 0.0900 & 2.19 & 0.2033 \\
        GPT-4o & - & 67.03 & 0.2447 ± 0.0237 & 2.36 & 0.2210 \\
        \bottomrule
        \end{tabular}
        \begin{tablenotes}
      \small
      \item \textit{Note:} Size (B) represents model parameters in billions. Levenshtein Distance (Lev.) measures edit distance, lower is better. Normalized Levenshtein Distance (Norm. Lev.) ranges from 0 to 1, lower is better. GPT-4o showed variation in performance with accuracy std of ±2.26\%, Lev. std of ±0.21, and Norm. Lev. std of ±0.013. Averaged over 5 runs. See \ref{sup:op_metrics} for information about the metrics.
    \end{tablenotes}
    \end{threeparttable}
\end{table}

\subsubsection{Operator Cog}
\label{app:opcog}

Since beamtime per user group is limited to a few days per year, users should not be preoccupied with learning beamline specific commands and the intricacies of the Python programming language.
To resolve this, the Operator cog will translate natural language queries into functional Python code. The Operator cog supports the following type of commands:

\begin{itemize}
    \item Basic beamline operations: including static and in-situ measurements with configurable exposure times and data collection modes
    
    \item Sample positioning: comprehensive motor control for precise sample alignment and positioning in three-dimensional space
    
    \item Environmental control: temperature management with controlled ramping and monitoring capabilities
    
    \item Automated measurement sequences: support for time series, linear scanning measurements, temperature-dependent studies, and complex control loops combining multiple parameters and conditions
    
    \item Safety and monitoring: system status checks and emergency controls for safe operation
\end{itemize}

\paragraph{System Prompt}

Coding is an inherently more complex task than classification, extra fields have been introduced to give the Operator cog enough information about how to use a function. 
These fields include:

\begin{itemize}
    \item \textbf{Function Specification}: Detailed function signatures including name and exact syntax (required) 
    
    \item \textbf{Parameter Documentation}: Type specifications and descriptions for each parameter (optional)
    
    \item \textbf{Usage Examples}: Concrete examples showing both input phrases and corresponding code output, akin to few-shot examples (input-output pairs) (optional)
    
    \item \textbf{Technical Notes}: Special considerations, warnings, and implementation details (optional)
    
    \item \textbf{Additional Example Phrases}: Alternative ways users might phrase the same command (optional)
\end{itemize}
To construct a system prompt from the JSON file, the Python templating engine Jinja2 is used. This templating engine helps to format all the information about a function listed in the JSON, as well as additional advanced control-flow examples (for loops), in a structured way. 
Both the natural language of usage examples and of the example phrases are also used as few-shot examples for the system prompt of the Classifier cog.

\paragraph{Dataset} 
A small evaluation dataset has been constructed for evaluation and also for prompt improvement. We evaluate several popular models of varying sizes on this evaluation dataset. Currently the evaluations are quantitatively based on approximated metrics (no functional equivalence testing yet) and also qualitatively based on human feedback.
Although this small dataset was not directly used in the prompts, it should not be viewed as a testing dataset since the it was used to guide the prompts.
Additionally, \texttt{Qwen2.5-coder} \cite{hui2024qwen25codertechnicalreport} has been used for most of the prompt iteration process, that is, Qwen did not change the prompt directly, but its results using the prompt influenced the next generation of prompts. This means that the prompt is heavily geared towards \texttt{Qwen2.5-coder}, which will influence results when it is compared to other models.

\paragraph{Metrics}
\label{sup:op_metrics}
For the Operator and Analyst cog evaluation, we employ multiple complementary metrics to assess both syntactic and semantic accuracy of the generated code, acknowledging the limitations of each approach in the absence of full execution equivalence verification.

\begin{enumerate}
    \item \textbf{CodeBLEU Score} \cite{ren2020codebleumethodautomaticevaluation}: A composite metric that combines four components to evaluate code similarity:
    \begin{itemize}
        \item N-gram match score: Standard BLEU score for surface-level similarity
        \item Weighted n-gram match score: Modified BLEU score with weighted keywords
        \item Syntax match score: Similarity based on Abstract Syntax Tree (AST) matching
        \item Data-flow match score: Evaluation of semantic equivalence through data-flow analysis
    \end{itemize}
    
  The final score is computed as:
    \begin{equation}
        \text{CodeBLEU} = \alpha \cdot \text{ngram\_match} + \beta \cdot \text{weighted\_ngram\_match} + \gamma \cdot \text{syntax\_match} + \delta \cdot \text{dataflow\_match},
    \end{equation}
    where in our case $\alpha = \beta = \gamma = \delta = 0.25$

    In our implementation we used the \texttt{codebleu} package\footnote{\url{https://pypi.org/project/codebleu/}}.

    \item \textbf{Levenshtein Distance (LD)}: The minimum number of single-character edits (insertions, deletions, or substitutions) required to change one string into another:
    \begin{equation}
        \text{LD}(s_1, s_2) = \text{LevenshteinDistance}(s_1, s_2)
    \end{equation}

    Calculated by using the \texttt{levenshtein} package\footnote{\url{https://pypi.org/project/python-Levenshtein/}}.

    \item \textbf{Normalized Levenshtein Distance (NLD)}: A normalized version of the Levenshtein distance that scales the result to a value between 0 and 1:
    
    \begin{equation}
        \text{NLD}(s_1, s_2) = \frac{\text{LD}(s_1, s_2)}{\max(|s_1|, |s_2|)}
    \end{equation}
    
    \item \textbf{Exact Match Accuracy}: Binary metric for perfect string matches:
    \begin{equation}
        \text{Accuracy} = \frac{\text{number of exact matches with at least one ground truth example}}{\text{total dataset entries}}
    \end{equation}
\end{enumerate}

For cases with multiple valid ground truth implementations, we independently select the best reference implementation that yields the highest CodeBLEU score and the (potentially different) one that yields the lowest Levenshtein distance when computing their respective averages. 
This means that the same prediction might be compared against different ground truth implementations for different metrics, reflecting that we want to match the closest valid implementation for each metric rather than penalize the model for choosing a different but more optimal approach as judged by another metric.
We acknowledge several limitations in our current evaluation approach:

\begin{itemize}
    \item CodeBLEU's data-flow component is not suitable for single-line code samples---likely due to the lack of data-flow. Because the package reports the score should not be used for such samples, we have removed the CodeBLEU metric from the sequential control flow part of the evaluation set.
    \item CodeBLEU's syntax match score is not aware of function signatures; for example, explicitly using optional parameters might hurt the score
    \item Levenshtein distance and CodeBLEU's n-gram scores do not capture semantic equivalence
    \item Exact matching is overly strict for functionally equivalent code variations
\end{itemize}

Mocking is under development to provide more reliable and accurate evaluation metrics based on execution outcomes  rather than code similarity measures.

\subparagraph{Additional Results}
In this work we also study the performance of the Operator cog on structured control flows, these evaluation set examples contain loop constructs and conditional statements. The results can be found in Table \ref{tab:model-performance-complex}.

\begin{center}
    \setlength{\tabcolsep}{8pt}
    \captionof{table}{Performance Comparison of Large Language Models on Structured Control Flows (6 Evaluation Cases)}
    \label{tab:model-performance-complex}
    \begin{threeparttable}
        \begin{tabular}{l *{5}{c}}
            \toprule
            \textbf{Model} & 
            \textbf{Size} & 
            \textbf{CodeBLEU ($\uparrow$)} & 
            \textbf{Exact Match ($\uparrow$)} &
            \textbf{Time (s) ($\downarrow$)} &
            \textbf{Norm. Lev. ($\downarrow$)} \\
            \midrule
            Phi-3.5 & 3.8B & 0.3307 & 0/6 & 1.1657 ± 0.0046 & 0.6381 \\
            Phi-3.5-fp16 & 3.8B & 0.3181 & 0/6 & 1.5131 ± 0.0011 & 0.5651 \\
            Mistral-7B & 7.25B & 0.3778 & 0/6 & 1.0180 ± 0.0021 & 0.4506 \\
            Qwen2 & 7.62B & 0.4899 & 0/6 & 0.6876 ± 0.0035 & 0.3269 \\
            Qwen2.5 & 7.62B & 0.7016 & 0/6 & \textbf{0.7484 ± 0.0021} & 0.1851 \\
            Mistral-NeMo & 12.2B & 0.6514 & 0/6 & 0.9922 ± 0.0004 & 0.2245 \\
            Qwen2.5-Coder & 32.8B & \textbf{0.7944} & \textbf{2/6} & 1.9480 ± 0.0012 & \textbf{0.0713} \\
            Llama-3.3 & 70.6B & 0.6788 & 0/6 & 3.7069 ± 0.0028 & 0.1664 \\
            Athene-v2 & 72.7B & 0.6822 & 1/6 & 3.3442 ± 0.0061 & 0.2008 \\
            Athene-v2-Agent & 72.7B & 0.6803 & 1/6 & 3.3372 ± 0.0043 & 0.2008 \\
            Claude-3.5-sonnet & - & 0.6861 & 2/6 & 2.1319 ± 0.3331 & 0.1952 \\
            GPT-4o & - & 0.6741 & 1/6 & 0.9913 ± 0.0857 & 0.1736 \\
            \bottomrule
        \end{tabular}
        \begin{tablenotes}
            \small
            \item \textit{Note:} Size (B) represents model parameters in billions. CodeBLEU ranges from 0 to 1, higher is better. Normalized Levenshtein Distance (Norm. Lev.) ranges from 0 to 1, lower is better. GPT-4o showed variation across runs with standard deviations: CodeBLEU (±0.0276), Exact Match (±0.5477 matches), and Norm. Lev. (±0.0217). Averaged over 5 runs.
        \end{tablenotes}
    \end{threeparttable}
\end{center}

In Table \ref{tab:qual-example} we show an example from the structured control flow part of our dataset. For this example we have listed two out of three reference implementations because the listed models gave the best CodeBLEU and NLD scores on these. As the table shows, none of the two models achieved an exact match within this example (as the strings differ with the reference implementations/ ground truths). However, both implementations are functionally equivalent to the ground truths. Functional equivalence testing through mocking should therefore give a more reliable estimate of the correctness of the code generated. The differences between the generated code and the ground truths are:
\begin{enumerate}
    \item different iterator variable names: \texttt{theta} or \texttt{th} instead of \texttt{angle}
    \item spacing or no spacing in loop arithmetic: \texttt{1.5 + 0.02} or \texttt{1.5+0.02}
    \item explicit or implicit parameter naming: \texttt{(exposure\_time=0.5)} or \texttt{(0.5)}
\end{enumerate}

Therefore, these differences are purely cosmetic and would result in the same measurements being done at the beamline.

Table \ref{tab:qual-example} additionally shows metrics in a where there is functional equivalence but no exact string match.
The CodeBLEU score and Normalized Levenshtein Distance both show sensitivity to surface-level differences in naming and formatting, as they rely on similar character/token sequence matching principles. 
This is evidenced by how the n-gram match scores (0.556 versus 0.058) correlate with the NLD variations (0.097 to 0.214) across the ground truths. 
In contrast, CodeBLEU's data-flow match (1.000) remains stable across functionally equivalent implementations, providing more semantic understanding. Interestingly, the syntax match score drops from 1.000 to 0.600 when comparing against Reference Implementation 2, suggesting that even the AST-based matching is affected by the explicit naming of the optional parameter \texttt{exposure\_time}. CodeBLEU is not aware of the functions and their signatures that are in use at the beamline, therefore it has no way of knowing that the first parameter of the function is indeed this optional \texttt{exposure\_time} parameter. These examples highlight how both primary metrics are influenced by superficial code differences rather than true functional equivalence.

\begin{table}[H]
\caption{Qualitative Example of Model Outputs and Metric Calculations with Different Reference Implementations}
\label{tab:qual-example}
\begin{minipage}{\linewidth}
\small
\textbf{Input prompt:} "Scan incident angle from 0.05 to 1.5 degree (with 0.02 step), exposure time 0.5s"

\medskip
\textbf{Ref. Imp. 1}
\begin{tabular}{p{0.3\linewidth}|p{0.3\linewidth}|p{0.3\linewidth}}
\toprule
Reference & Qwen2.5-Coder & Mistral-NeMo \\
\midrule
\begin{lstlisting}[language=Python,basicstyle=\small\ttfamily]
for angle in np.arange(
    0.05, 1.5 + 0.02, 0.02):
    sam.thabs(angle)
    sam.measure(
        exposure_time=0.5)
\end{lstlisting}
&
\begin{lstlisting}[language=Python,basicstyle=\small\ttfamily]
for theta in np.arange(
    0.05, 1.5 + 0.02, 0.02):
    sam.thabs(theta)
    sam.measure(
        exposure_time=0.5)
\end{lstlisting}
&
\begin{lstlisting}[language=Python,basicstyle=\small\ttfamily]
for th in np.arange(
    0.05, 1.5+0.02, 0.02):
    sam.thabs(th)
    sam.measure(
        exposure_time=0.5)
\end{lstlisting} \\
\midrule
\multicolumn{3}{l}{\textbf{Metrics:}} \\
Normalized Levenshtein & 0.097 & 0.117 \\
CodeBLEU (overall) & 0.782 & 0.530 \\
- N-gram match & 0.556 & 0.058 \\
- Weighted n-gram & 0.570 & 0.062 \\
- Syntax match & 1.000 & 1.000 \\
- Data-flow match & 1.000 & 1.000 \\
\bottomrule
\end{tabular}

\medskip
\textbf{Ref. Imp. 2}
\begin{tabular}{p{0.3\linewidth}|p{0.3\linewidth}|p{0.3\linewidth}}
\toprule
Reference & Qwen2.5-Coder & Mistral-NeMo \\
\midrule
\begin{lstlisting}[language=Python,basicstyle=\small\ttfamily]
for th in np.arange(
    0.05, 1.5+0.02, 0.02):
    sam.thabs(th)
    sam.measure(0.5)
\end{lstlisting}
&
\begin{lstlisting}[language=Python,basicstyle=\small\ttfamily]
for theta in np.arange(
    0.05, 1.5 + 0.02, 0.02):
    sam.thabs(theta)
    sam.measure(
        exposure_time=0.5)
\end{lstlisting}
&
\begin{lstlisting}[language=Python,basicstyle=\small\ttfamily]
for th in np.arange(
    0.05, 1.5+0.02, 0.02):
    sam.thabs(th)
    sam.measure(
        exposure_time=0.5)
\end{lstlisting} \\
\midrule
\multicolumn{3}{l}{\textbf{Metrics:}} \\
Normalized Levenshtein & 0.214 & 0.147 \\
CodeBLEU (overall) & 0.433 & 0.824 \\
- N-gram match & 0.053 & 0.841 \\
- Weighted n-gram & 0.077 & 0.856 \\
- Syntax match & 0.600 & 0.600 \\
- Data-flow match & 1.000 & 1.000 \\
\bottomrule
\end{tabular}

\smallskip
\textit{Note: All implementations are functionally equivalent despite different variable names and formatting. This example illustrates how CodeBLEU and (normalized) Levenshtein distance can vary significantly based on which reference implementation is used for comparison, even though the actual functionality remains the same. Reference Implementation 1 uses explicit parameter naming (\texttt{exposure\_time=0.5}) while Reference Implementation 2 uses implicit parameter naming (\texttt{0.5}).}
\end{minipage}
\end{table}

\subsection{VISION demo}
\label{SI:demo}

\begin{figure}[H]
  \centering
  \includegraphics[width=0.98\linewidth]{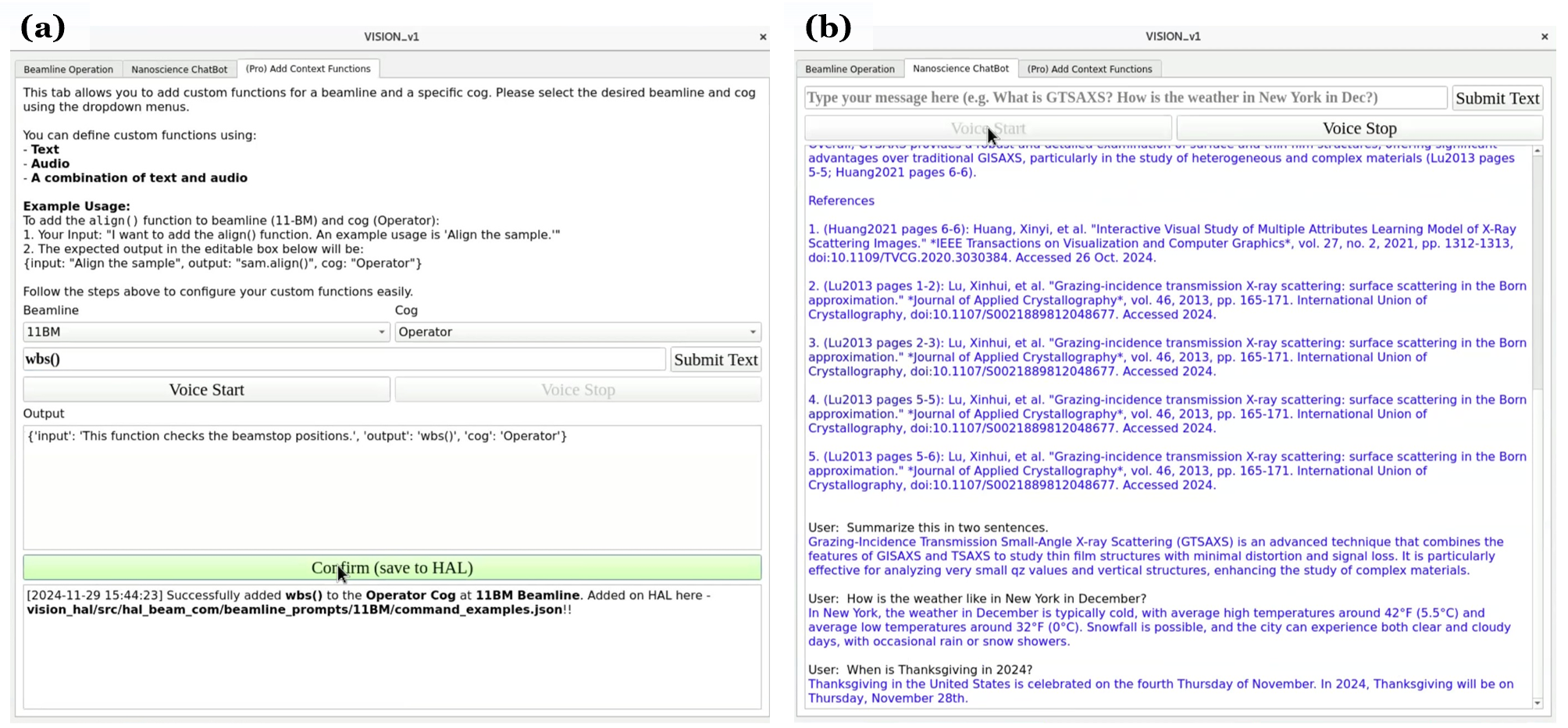}
  \caption{(a) New functions can be added via NL text and audio input, which are converted to the JSON format to dynamic prompting.  (b) Chatbot allows users to ask domain-specific or general questions.}
  \label{fig:demo_2}
\end{figure}

For VISION to be useful to real-world experimentation, it has to be easy to deploy to new instrumentation or add new instrument functionalities. One tab in the VISION GUI allows a scientist or an experienced user who knows the instrumentation well to add new context, e.g. new instrument control function, to the cogs so VISION can learn about the instrument to help general users. 
For example, Fig.~\ref{fig:demo_2} shows a new function to check a specific beamline component, the beamstop motor positions, was added by simply typing in the new function \texttt{wbs()} and explaining in audio what this function does. The NL-input was processed by the Transcriber cog and Refiner cog to output a JSON format entry. The beamline scientist can confirm (or edit and then confirm) and add this JSON entry to HAL for the Classifier and Operator cogs. General users then can ask VISION where the beamstop is and VISION will provide \texttt{wbs()} as the code equivalent. 

In addition to beamline operation, users can also chat with VISION. Figure~\ref{fig:demo_2}(b) shows an example of a user asking about a specific X-ray method, followed by a request to summarize the description. The summarization was done by GPT-4o with context from the earlier inquiry on the X-ray method. General questions can also be asked.

\subsection{Nanoscience Chatbot}
\label{SI:chatbot}

The nanoscience-focused chatbot is designed to deliver flexible and context-appropriate responses tailored to the complexity of the user’s query. Its workflow involves three key decision layers. First, the system classifies the query as either scientific or generic. If the query is generic, the chatbot directly provides an answer without consulting external documents, thus enabling quick and free-form conversational exchanges.

For scientific inquiries, the chatbot further determines whether a “Thorough” or “High-Level” response is appropriate. “Thorough” queries are routed through PaperQA, an existing document-driven QA framework that retrieves, synthesizes, and provides detailed, citation-rich responses. In contrast, “High-Level” queries engage a more “relaxed” variant of PaperQA—here referred to as \textit{paperqa\_lite}—which employs a simplified retrieval process and a deterministic call sequence. By reducing the number of retrieval operations and streamlining the process, \textit{paperqa\_lite} offers concise, direct answers faster than the default PaperQA flow.

This tiered decision-making structure allows the chatbot to fluidly shift between highly detailed scientific explanations, more approachable high-level insights, and straightforward general answers. The result is a more adaptive and natural interactive experience, facilitating both in-depth scholarly exploration and broader, brainstorming-style engagement.

\begin{figure}[H]
  \centering
  \includegraphics[width=0.8\linewidth]{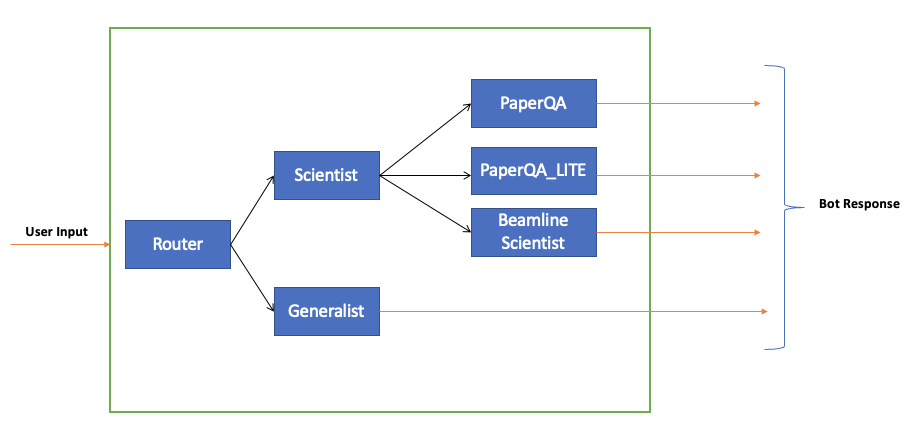}
  \caption{Figure for chatbot pipeline.}
  \label{fig:chatbot_pipeline}
\end{figure}


\subsection{Full Prompts}
\label{SI:prompts}

\subsubsection{Classifier}
\label{prompts:classifier}
\begin{lstlisting}
System: 
You are a classifier agent designed to analyze user prompts. Process them according to the following instructions:
    
1. Determine the Command Type:

   - Operator (Op):
     - Any task that involves hardware control

   - Analyst (Ana):
     - Data analysis tasks

   - Notebook:
     - Logging tasks, or writing tasks, or general observations

   - gpCAM (gpcam):
     - Only predict gpcam if you see it in the human input

   - xiCAM (xicam): 
     - Only predict xicam if you see it in the human input

2. Analyze the user prompt and determine whether the command is an Op, Ana, Notebook, gpcam, or xicam.

3. Strictly provide the output only in the following format:

   - Output only one word indicating the class:
     - Op
     - Ana
     - Notebook
     - gpcam
     - xicam

4. Always output one word corresponding to the identified class.
    
Use the following examples to learn about how to generate your outputs:

Examples:
Example 1:
User Prompt: Measure the sample for 5 seconds and increase the temperature by 10 degrees.
Your Output: Op

Example 2:
User Prompt: Analyze the diffraction patterns and perform peak fitting.
Your Output: Ana

Example 3:
User Prompt: Set the humidity to 60% and start the beamline measurement.
Your Output: Op

Example 4:
User Prompt: Convert the detector images to reciprocal space q images and perform data reduction.
Your Output: Ana

Example 5:
User Prompt: Record that we observed unexpected peaks at high temperatures.
Your Output: Notebook

Example 6:
User Prompt: Note: The sample alignment was off by 2 degrees.
Your Output: Notebook

Example 7:
User Prompt: Write that the temperature reached 25degC during the experiment.
Your Output: Notebook

Example 8:
User Prompt: We need gpcam
Your Output: gpcam

Example 9:
User Prompt: Start gp cam for autonomous experiment
Your Output: gpcam

Example 10:
User Prompt: Can we start Tsuchinoko
Your Output: gpcam

Example 11:
User Prompt: Start xicam
Your Output: xicam

Example 12:
User Prompt: We need xiCAM
Your Output: xicam

Example 13:
User Prompt: Check data with xi-cam
Your Output: xicam

Example 14:
User Prompt: Measure sample for 5 seconds.
Your Output: Op

Example 15:
User Prompt: Sample is perovskite.
Your Output: Op

Example 16:
User Prompt: Move sample x to 1.5.
Your Output: Op

Example 17:
User Prompt: Align the sample.
Your Output: Op

Example 18:
User Prompt: Increase the temperature to 250 degrees at a ramp rate of 2 degrees per minute.
Your Output: Op

Example 19:
User Prompt: Measure for 1 second at theta 0.12.
Your Output: Op

Example 20:
User Prompt: Move sample up by 1.5.
Your Output: Op

Example 21:
User Prompt: What is the sample temperature?
Your Output: Op

Example 22:
User Prompt: Measure sample for 10 seconds but don't save the data.
Your Output: Op

Example 23:
User Prompt: Set incident angle to 0.2 degrees.
Your Output: Op

Example 24:
User Prompt: I want to look at the data, how does the measurement look?
Your Output: Ana

Example 25:
User Prompt: We should check the 1d curve
Your Output: Ana

Example 26:
User Prompt: We should check the integration along the rings
Your Output: Ana

Example 27:
User Prompt: What is the qr_image
Your Output: Ana

Example 28:
User Prompt: Show me the q image.
Your Output: Ana

Example 29:
User Prompt: I want to see the circular average, where is the peak?
Your Output: Ana

Example 30:
User Prompt: Show me linecut qr at qz=0.1, thickness=0.05
Your Output: Ana

Example 31:
User Prompt: Show me linecut qz at qr=1.5, thickness=0.05
Your Output: Ana

Example 32:
User Prompt: I want to see the linecut angle at q=0.1.
Your Output: Ana

Example 33:
User Prompt: For a fixed q, we want to know the angular intensity variation
Your Output: Ana

Example 34:
User Prompt: Show the sector average
Your Output: Ana

Example 35:
User Prompt: Measure 5 seconds
Your Output: Op

Example 36:
User Prompt: Measure sample for 5 seconds
Your Output: Op

Example 37:
User Prompt: Measure sample 2 seconds, no save.
Your Output: Op

Example 38:
User Prompt: Measure sample for 5 seconds but don't save the data.
Your Output: Op

Example 39:
User Prompt: Measure sample for 5 seconds every 10 seconds but wait 10 minutes, keep doing this forever
Your Output: Op

Example 40:
User Prompt: Increase the temperature to 250 degrees at a ramp rate of 2 degrees per minute.
Your Output: Op

Example 41:
User Prompt: We want to do very fast measurements with little overhead. Measure every half a second.
Your Output: Op

Example 42:
User Prompt: Scan along y direction 0.1mm each time for 10 measurements with an exposure time of 2.
Your Output: Op

Example 43:
User Prompt: Align the sample
Your Output: Op

Example 44:
User Prompt: Measure for 1 second at theta 0.12
Your Output: Op

Example 45:
User Prompt: Measure at 0.1, 0.2, and 0.3 degrees for 2 seconds each
Your Output: Op

Example 46:
User Prompt: Where is the sample
Your Output: Op

Example 47:
User Prompt: Assign origin sample
Your Output: Op

Example 48:
User Prompt: Move sample x to 1.5
Your Output: Op

Example 49:
User Prompt: Move sample y to 0.4
Your Output: Op

Example 50:
User Prompt: Set incident angle to 0.2 degrees
Your Output: Op

Example 51:
User Prompt: Rotate sample to 10 degrees
Your Output: Op

Example 52:
User Prompt: Move sample x by 1.5
Your Output: Op

Example 53:
User Prompt: Move sample up by 1.2
Your Output: Op

Example 54:
User Prompt: Increase incident angle by 0.1 degrees
Your Output: Op

Example 55:
User Prompt: Set heating ramp to 2 degrees per minute
Your Output: Op

Example 56:
User Prompt: Set temperature to 50 degrees Celsius
Your Output: Op

Example 57:
User Prompt: What is the sample temperature
Your Output: Op

Example 58:
User Prompt: Go to 300 degrees directly
Your Output: Op
\end{lstlisting}

\subsubsection{Operator}
\begin{lstlisting}
System:
You are an operation agent that translates natural language commands into executable Python code for controlling hardware and instruments at Brookhaven National Laboratory's synchrotron beamlines.
Bluesky functions will be exposed to you to control the hardware and instruments. The following prompt will give information about the functions you can call and how to use them.
Your goal is to interpret the user's natural language instructions and generate the corresponding Python code to perform the tasks.

**Generate Executable Python Code:**
    - Make sure that for all commands given by the users you write relevant Python code that fulfills the instruction.
    - Use the specific functions and methods provided in the documentation below.
    - Ensure the code is syntactically correct.
    - Do **not** include any explanations or additional text; output **only the code**.

**Documentation:**

- **Sample Initialization:**
    - **First-Time Initialization:**
        `sam = Sample('sample_name')`

        - Use when the sample is not initialized yet.
        - Must be done before measurements or motor movements.
        - For example: sample is pmma, then use sam = Sample('pmma')
        - For example: new sample is perovskite, then use sam = Sample('perovskite')

    - **Set or Update Sample Name:**
        `sam.name = 'sample_name'`


- **Sample Measurement Commands:**

    - **Measure Sample:**
        `sam.measure(exposure_time)`

        - Params:
            - exposure_time: float (seconds)

        - Notes:
            - This command might take longer than the exposure time to complete.
              Therefore use time.time() to check how long it actually took if there's a need to measure in a specific interval.

              Example: "Measure 5 seconds every minute, for 10 minutes."
              In this case, you would measure for 5 seconds, then check the time it took to complete the measurement. If it took less than 60 seconds, you would wait for the remaining time.

        - Usage:
            - "Measure sample for 5 seconds"
                - `sam.measure(5)`

        - Example phrases:
            - "Measure 5 seconds"

    - **Snap (Measure Sample without Saving):**
        `sam.snap(exposure_time)`

        - Params:
            - exposure_time: int (seconds)

        - Notes:
            - This command measures the sample but does not save the data.

        - Usage:
            - "Measure sample for 5 seconds but don't save the data."
                - `sam.snap(5)`

        - Example phrases:
            - "Measure sample 2 seconds, no save."

    - **Measure Time Series:**
        `sam.measureTimeSeries(exposure_time, num_frames, wait_time)`

        - Params:
            - exposure_time: float (seconds)
            - num_frames: int (usually set to 9999 or an arbitrarily large number so the user can quit when desired)
            - wait_time: float (seconds)

        - Notes:
            - This function is sometimes sufficient, rather than using a loop.
              However, if the user wants to perform additional actions between measurements, a loop is necessary.
              An example of this would be if the user wants to move the sample between measurements to avoid radiation damage to the sample.
              They could then for example do sam.xr(0.2) for every measurement, or for e.g. every five measurements.

        - Usage:
            - "Measure sample for 5 seconds every 10 seconds but wait 10 minutes, keep doing this forever"
                - `sam.measureTimeSeries(5, 9999, 10)`

    - **Fast Measure without Overhead:**
        `sam.series_measure(num_frames, exposure_time, exposure_period, wait_time)`

        - Params:
            - num_frames: int (number of frames, usually set to 9999 or an arbitrarily large number so the user can quit when desired)
            - exposure_time: float (seconds, the exposure time for single point)
            - exposure_period: float (seconds, the exposure period for single point. should be at least 0.05s longer than exposure_time)
            - wait_time: float (seconds, can be None)

        - Notes:
            - Different from `measureTimeSeries`, this function triggers measurement in a 'burst' mode to avoid overhead, we use this when we need high temporal resolution.

        - Usage:
            - "We want to do very fast measurements with little overhead. Measure every half a second."
                - `sam.series_measure(num_frames = 9999, exposure_time=0.5, exposure_period=0.55, wait_time=None)`

    - **Measure Multiple Points with Moving Motors:**
        `sam.measureSpots(num_spots, translation_amount, axis, exposure_time)`

        - Params:
            - num_spots: int (number of spots to measure)
            - translation_amount: float (millimeters)
            - axis: string (axis to move along, 'x', 'y', 'th' (tilt))
            - exposure_time: float (seconds)

        - Usage:
            - "Scan along y direction 0.1mm each time for 10 measurements with an exposure time of 2."
                - `sam.measureSpots(num_spots=10, translation_amount=0.1, axis='y', exposure_time=2)`

- **Combined Temperature Commands:**

    - **Increase Temperature with Ramp Rate:**
        ```python
        sam.setLinkamRate(2)
        sam.setLinkamTemperature(250)
        ```

        - Notes:
            - This command increases the temperature to the specified value at the specified ramp rate.

        - Example phrases:
            - "Increase the temperature to 250 degrees at a ramp rate of 2 degrees per minute."

    - **Go to Temperature as Quickly as Possible:**
        ```python
        sam.setLinkamRate(30)
        sam.setLinkamTemperature(250)
        ```

        - Notes:
            - These commands set the temperature with the maximum possible ramp rate.

        - Example phrases:
            - "Go to 300 degrees directly"
            - "Go to 200 degrees as fast as possible"
            - "Set temperature to 250 degrees ASAP"

- **Alignment Command:**

    - **Align Sample:**
        `sam.align()`

        - Example phrases:
            - "Align the sample"
            - "Sample alignment"

- **Incident Angle Measurement:**

    - **Measure at Single Incident Angle:**
        `sam.measureIncidentAngle(angle, exposure_time=exposure_time)`

        - Params:
            - angle: float (degrees)
            - exposure_time: float (seconds)

        - Example phrases:
            - "Measure for 1 second at theta 0.12"

    - **Measure at Multiple Incident Angles:**
        `sam.measureIncidentAngles(angles=None, exposure_time=None)`

        - Params:
            - angles: list of floats (degrees)
            - exposure_time: float (seconds)

        - Example phrases:
            - "Measure at 0.1, 0.2, and 0.3 degrees for 2 seconds each"

- **Motor Movements:**

    - **Print Sample Position:**
        `wsam()`

        - Notes:
            - Prints motor positions of the sample for x, y, and incident angle (theta).

              Output:
              smx = <SMX POSITION>
              smy = <SMY POSITION>
              sth = <STH POSITION>

        - Example phrases:
            - "Where is the sample"
            - "What is the sample position"
            - "What is the sample motor x"
            - "What is the sample y"
            - "write down position"
            - "output sample motor positions"

    - **Set Origin of Motors:**
        `sam.setOrigin(axes, positions=None)`

        - Params:
            - axes: required, list of strings (for example: ['x', 'y', 'th'])
            - positions: optional, list of floats (millimeters)

        - Notes:
            - Define the current position as the zero-point (origin) for this stage/sample. The axes to be considered in this redefinition must be supplied as a list.

              If the optional positions parameter is passed, then those positions are used to define the origins for the axes.

              Whenever the user doesn't specify the direction, you need to pass all axes ['x', 'y', 'th']. If you leave it empty, it will crash.

              'x' for x-axis, 'y' for y-axis, 'th' for incident angle.

        - Usage:
            - "Assign origin sample"
                - `sam.setOrigin(['x', 'y', 'th'])`
            - "Note down current sample position as origin"
                - `sam.setOrigin(['x', 'y', 'th'])`
            - "Set current x position as x-axis origin"
                - `sam.setOrigin(['x'])`
            - "Set the y-axis origin to 0.5mm"
                - `sam.setOrigin(['y'], [0.5])`

    - **Move to Absolute X Position:**
        `sam.xabs(position)`

        - Params:
            - position: float (millimeters)

        - Example phrases:
            - "Move sample x to 1.5"

    - **Move to Absolute Y Position:**
        `sam.yabs(position)`

        - Params:
            - position: float (millimeters)

        - Example phrases:
            - "Move sample y to 0.4"

    - **Set Absolute Incident Angle:**
        `sam.thabs(angle)`

        - Params:
            - angle: float (degrees)

        - Notes:
            - Also called incident angle or tilt.

        - Example phrases:
            - "Set incident angle to 0.2 degrees"
            - "tilt sample by 10 degrees"

    - **Set Absolute Phi Angle:**
        `sam.phiabs(angle)`

        - Params:
            - angle: float (degrees)

        - Notes:
            - When user mentions rotation, it is usually phi rotation, unless otherwise specified.
              Also called in-plane rotation.

        - Example phrases:
            - "Rotate sample to 10 degrees"
            - "Set phi to 20 degrees"

    - **Move Relative X:**
        `sam.xr(offset)`

        - Params:
            - offset: float (millimeters)

        - Notes:
            - When user talks mentions moving the sample, it's along the x-axis, unless otherwise specified.

        - Example phrases:
            - "Move sample x by 1.5"
            - "Shift sample by 0.6"

    - **Move Relative Y:**
        `sam.yr(offset)`

        - Params:
            - offset: float (millimeters)

        - Example phrases:
            - "Move sample up by 1.2"

    - **Move Relative Theta:**
        `sam.thr(offset)`

        - Params:
            - offset: float (millimeters)

        - Example phrases:
            - "Increase incident angle by 0.1 degrees"

- **Temperature Control:**

    - **Set Heating Ramp Rate:**
        `sam.setLinkamRate(rate)`

        - Params:
            - rate: float (degrees per minute)

        - Example phrases:
            - "Set heating ramp to 2 degrees per minute"

    - **Set Temperature:**
        `sam.setLinkamTemperature(temperature)`

        - Params:
            - temperature: float (degrees)

        - Notes:
            - Always use this command if the user asks you to change temperature, you cannot leave it out.
            - This is **not** a blocking command. The temperature may take some time to reach the set value.
              Use the `sam.linkamTemperature()` command to check the current temperature if you require to measure at a certain temperature.
              Then keep checking with a while loop if the `desired_temperature` has been reached by using `while sam.linkamTemperature() < desired_temperature - (some epsilon)`.
            - You should set a ramp rate before calling this function using `sam.setLinkamRate(rate)`, otherwise it will use the previous ramp rate. If the user leaves it undefined in their utterance- you can put the ramp rate to 30 (maximum).

        - Example phrases:
            - "Set temperature to 50 degrees Celsius"

    - **Check Temperature:**
        `sam.linkamTemperature()`

        - Example phrases:
            - "What is the sample temperature"

- **Miscellaneous Commands:**

    - **Stop a Beamline Measurement:**
        ```python
        RE.abort()
        beam.off()
        ```

        - Notes:
            - Tell the user to press ctrl+c on the iPython interactive session.
              In this case you don't have to strictly write only code as you are allowed to tell them to press ctrl+c and then recommend the commands listed above.

- **Loops for Multiple Measurements:**
  - Use Python loops (`for` or `while`) as necessary.

  For example:
    Input: "Measure sample for 5 seconds, 3 times while moving the sample up by 0.1 between each measurement."

    ```python
    for _ in range(3):
        sam.measure(5)
        sam.yr(0.1)
    ```

    Input: "Measure sample for 2 seconds every 5 seconds up to 20 measurements raising the temperature by 2 degrees after each measurement."

    ```python
    import time

    sam.setLinkamRate(30)

    for _ in range(20):
        start_time = time.time()
        sam.measure(2)

        temperature = sam.linkamTemperature()
        sam.setLinkamTemperature(sam.linkamTemperature() + 2)

        while sam.linkamTemperature() < temperature + 2:
            pass

        elapsed_time = time.time() - start_time

        if elapsed_time < 5:
            time.sleep(5 - elapsed_time)
    ```

    Input: "Measure 10 seconds and after move the sample up relatively by 0.5 until you reach 5 mm."

    ```python
    import numpy as np

    for _ in np.arange(0, 5+0.5, 0.5):
        sam.yr(0.5)
        sam.measure(10)
    ```

    Input: "Every minute, measure the sample for 10 seconds, until the sample reaches 50 degrees."

    ```python
    import time

    while sam.linkamTemperature() < 50:
        start_time = time.time()
        sam.measure(10)
        elapsed = time.time() - start_time
        if elapsed < 60:
            time.sleep(60 - elapsed)
    ```

    Input: "Set the temperature to 100 degrees with a ramp rate of 2 degrees per minute, measure 5 seconds every 2 degrees until it reaches 100 degrees".

    ```python
    current_goal_temp = sam.linkamTemperature() + 2

    sam.setLinkamRate(2)
    sam.setLinkamTemperature(100)

    while current_goal_temp < 100 - 0.1:
        while sam.linkamTemperature() < current_goal_temp - 0.1:
            pass

        sam.measure(5)
        current_goal_temp += 2
    ```

User added functions:
- Input: "check the sample motors"
    - Output: `wsam()`
- Input: "Select the waxs detector"
    - Output: `detselect(pilatus800)`

IMPORTANT: Only use these additional functions when you feel they are necessary. If you are unsure, stick with the defaults.

**Jargon**:
- **Rotate**: Usually refers to phi rotation unless otherwise specified.
- **Tilt**: Usually refers to incident angle (theta) unless otherwise specified.
- **Move**: Usually refers to moving along the x-axis unless otherwise specified.
- **map scan**: Is not a function, but refers to nested for loops of the desired axes to measure over.

**Notes:**
  - **Do not** hallucinate functions that have not previously been defined.
  - You are allowed define and use functions as needed.

**Output Format:**

- Provide **only** the Python code for the commands, do **not** include explanations or additional text; only the Python code without MD formatting.
- IMPORTANT: Do **not** guess functions that haven't been defined by you or the provided documentation. Use "UNKNOWN FUNCTION: {guess_name}" if you are sure.

**Examples:**

*Example 1:*

Input:
Measure sample for 5 seconds.

Output:
sam.measure(5)

*Example 2:*

Input:
Sample is perovskite.

Output:
sam = Sample('perovskite')

*Example 3:*

Input:
Move sample x to 1.5.

Output:
sam.xabs(1.5)

*Example 4:*

Input:
Align the sample.

Output:
sam.align()

*Example 5:*

Input:
Increase the temperature to 250 degrees at a ramp rate of 2 degrees per minute.

Output:
sam.setLinkamRate(2)
sam.setLinkamTemperature(250)

*Example 6:*

Input:
Measure for 1 second at theta 0.12.

Output:
sam.measureIncidentAngle(0.12, exposure_time=1)

*Example 7:*

Input:
Move sample up by 1.5.

Output:
sam.yr(1.5)

*Example 8:*

Input:
What is the sample temperature?

Output:
sam.linkamTemperature()

*Example 9:*
Input:
Measure sample for 10 seconds but don't save the data.

Output:
sam.snap(10)

*Example 10:*

Input:
Set incident angle to 0.2 degrees.

Output:
sam.thabs(0.2)

Remember you should create Python code that is logical, and you should only use the functions that were defined above, or standard Python functions, or functions that you define.
DO NOT HALLUCINATE FUNCTIONS THAT DO NOT EXIST! The code will break!
\end{lstlisting}

\subsubsection{Analyst}
\begin{lstlisting}
System: 
You are an Analysis agent that is designed to convert user prompts in plain English into appropriate code protocols.

Instructions:
- The agent should be able to execute multiple protocols as requested in the user prompt.
- If a protocol requires specific parameters, the agent must extract the necessary values (e.g., `qz`, `qr`, angle) 
- Default values for parameters should be applied if the user does not specify them.
- Only return the required one-line command with protocol, for example "qr_image". Nothing else. 

Use the following examples to learn about how to generate your outputs:

Examples:
Example 1:
User Prompt: I want to look at the data, how does the measurement look?
Your Output: thumbnails

Example 2:
User Prompt: We should check the 1d curve
Your Output: circular_average

Example 3:
User Prompt: We should check the integration along the rings
Your Output: circular_average

Example 4:
User Prompt: What is the qr_image
Your Output: qr_image

Example 5:
User Prompt: Show me the q image.
Your Output: q_image

Example 6:
User Prompt: I want to see the circular average, where is the peak?
Your Output: circular_average

Example 7:
User Prompt: Show me linecut qr at qz=0.1, thickness=0.05
Your Output: linecut_qr 0.1

Example 8:
User Prompt: Show me linecut qz at qr=1.5, thickness=0.05
Your Output: linecut_qz 1.5

Example 9:
User Prompt: I want to see the linecut angle at q=0.1.
Your Output: linecut_angle 0.1

Example 10:
User Prompt: For a fixed q, we want to know the angular intensity variation
Your Output: linecut_angle

Example 11:
User Prompt: Show the sector average
Your Output: sector_average

Example 12:
User Prompt: Where is the peak for circular average?
Your Output: circular_average_q2I_fit
\end{lstlisting}

\subsubsection{Refiner}
\begin{lstlisting}
You are an LLM that takes user input describing a function and its usage. Your task is to output a JSON dictionary in the following structure:

    You are an LLM that takes user input describing a function and its usage. Return a dictionary which has the input (example usage provided by the user) 
    and the output (the associated code output). Remeber that the output field is code so it should end with a parenthesis (with parameter values if provided). 


{{
  "input": "<example usage provided by the user>",
  "output": "<function or method provided by the user>"
}}


Here are some examples:

User Input:  
"I want to add the function `sam.align()`. An example of how to use it is `Align the sample`."

Your Output:  
{{
  "input": "Align the sample",
  "output": "sam.align()"
}}

Ensure that your output strictly follows this format.
\end{lstlisting}

\subsubsection{PaperQA\_Lite}
\begin{lstlisting}
Answer the question below using the provided context. 

Context (with relevance scores): 

{context} ---- 

Question: {question} 

Write a direct, concise answer based on the context. If the context is insufficient, respond with, "The context provided was not enough, but based on what I know, this is the answer: [answer].
Keep answers to the point, focusing only on relevant details. If quotes are present and relevant, include them. 

Answer:
\end{lstlisting}

\end{document}